\documentclass[11pt]{article}

\usepackage{acl}

\usepackage{times}
\usepackage{latexsym}
\usepackage{booktabs}
\usepackage{xcolor}
\usepackage{multirow}
\usepackage[T1]{fontenc}
\usepackage[utf8]{inputenc}
\usepackage{microtype}
\usepackage{graphicx}
\usepackage{amsmath}
\usepackage{amsfonts}
\usepackage{array}
\usepackage{algorithm}
\usepackage{algorithmic}
\usepackage{wasysym}
\usepackage{multirow}
\usepackage{enumitem}
\usepackage[switch]{lineno}
\usepackage{tabularx}
\usepackage[normalem]{ulem}
\usepackage[para]{footmisc}
\usepackage[english]{babel}
\usepackage{pdfrender}
\usepackage{tikz}
\usepackage{subfigure}
\usepackage{scalerel,graphicx,xparse}
\usepackage{amssymb}% http://ctan.org/pkg/amssymb
\usepackage{pifont}% http://ctan.org/pkg/pifont
\newcommand{\xmark}{\ding{55}}%
\NewDocumentCommand\poopemoji{}{\scalerel*{\includegraphics{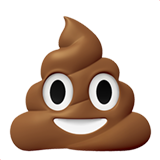}}{X}}
\makeatletter
\newcommand*\bigcdot{\mathpalette\bigcdot@{.5}}
\newcommand*\bigcdot@[2]{\mathbin{\vcenter{\hbox{\scalebox{#2}{$\m@th#1\bullet$}}}}}
\makeatother

\newcommand\bigforall{\mbox{\Large $\mathsurround0pt\forall$}} 

\newcolumntype{H}{>{\setbox0=\hbox\bgroup}c<{\egroup}@{}}

\newcommand{\mymethod}{\textsc{Anthro}}
\newcommand{\mymethodx}{$\mathrm{\textsc{Anthro}}_\beta$}

\title{Perturbations in the Wild: Leveraging Human-Written Text Perturbations for Realistic Adversarial Attack and Defense}

\usepackage{tikz}
\usepackage{lipsum}

\newcommand\copyrighttext{%
  \footnotesize \textcolor{red}{\textbf{DISCLAIMER! THIS PAPER CONTAINS EXAMPLE TEXTS THAT ARE OFFENSIVE IN NATURE}}}
\newcommand\copyrightnotice{%
\begin{tikzpicture}[remember picture,overlay]
\node[anchor=south,yshift=18pt] at (current page.south) {{\copyrighttext}};
\end{tikzpicture}%
}

\author{Thai Le\hspace{0.2in} Jooyoung Lee\hspace{0.2in} Kevin Yen$^*$\hspace{0.2in} Yifan Hu$^*$\hspace{0.2in} Dongwon Lee \vspace{0.1in}\\
        \begin{tabular}{cc}
          Penn State University   &          \{thaile, jfl5838, dongwon\}@psu.edu\\
           Yahoo Research$^*$  & \{kevinyen, yifanhu\}@yahooinc.com$^*$
        \end{tabular}
}
  
\begin{document}

\maketitle
\copyrightnotice
\begin{abstract}

We proposes a novel algorithm, {\mymethod},  that {\em inductively} extracts over 600K human-written text perturbations in the wild and leverages them for {\em realistic} adversarial attack. Unlike existing character-based attacks which often deductively hypothesize a set of manipulation strategies, our work is grounded on actual observations from real-world texts. We find that adversarial texts generated by {\mymethod} achieve the best trade-off between (1) attack success rate, (2) semantic preservation of the original text, and (3) stealthiness--i.e. indistinguishable from human writings hence harder to be flagged as suspicious. Specifically, our attacks accomplished around 83\% and 91\% attack success rates on BERT and RoBERTa, respectively. Moreover, it outperformed the \textit{TextBugger} baseline with an increase of 50\% and 40\% in terms of semantic preservation and stealthiness when evaluated by both layperson and professional human workers. {\mymethod} can further enhance a BERT classifier's performance in understanding different variations of human-written toxic texts via adversarial training when compared to the Perspective API. \textit{Source code will be published at \url{github.com/lethaiq/perturbations-in-the-wild}}.

\end{abstract}

\section{Introduction}\label{sec:introduction}

\begin{figure}[t!]
% https://app.diagrams.net/#G1l9fJ6BC3A8MPywE2m6hxkYrXsVI4AqUr
\centerline{\includegraphics[width=0.47\textwidth]{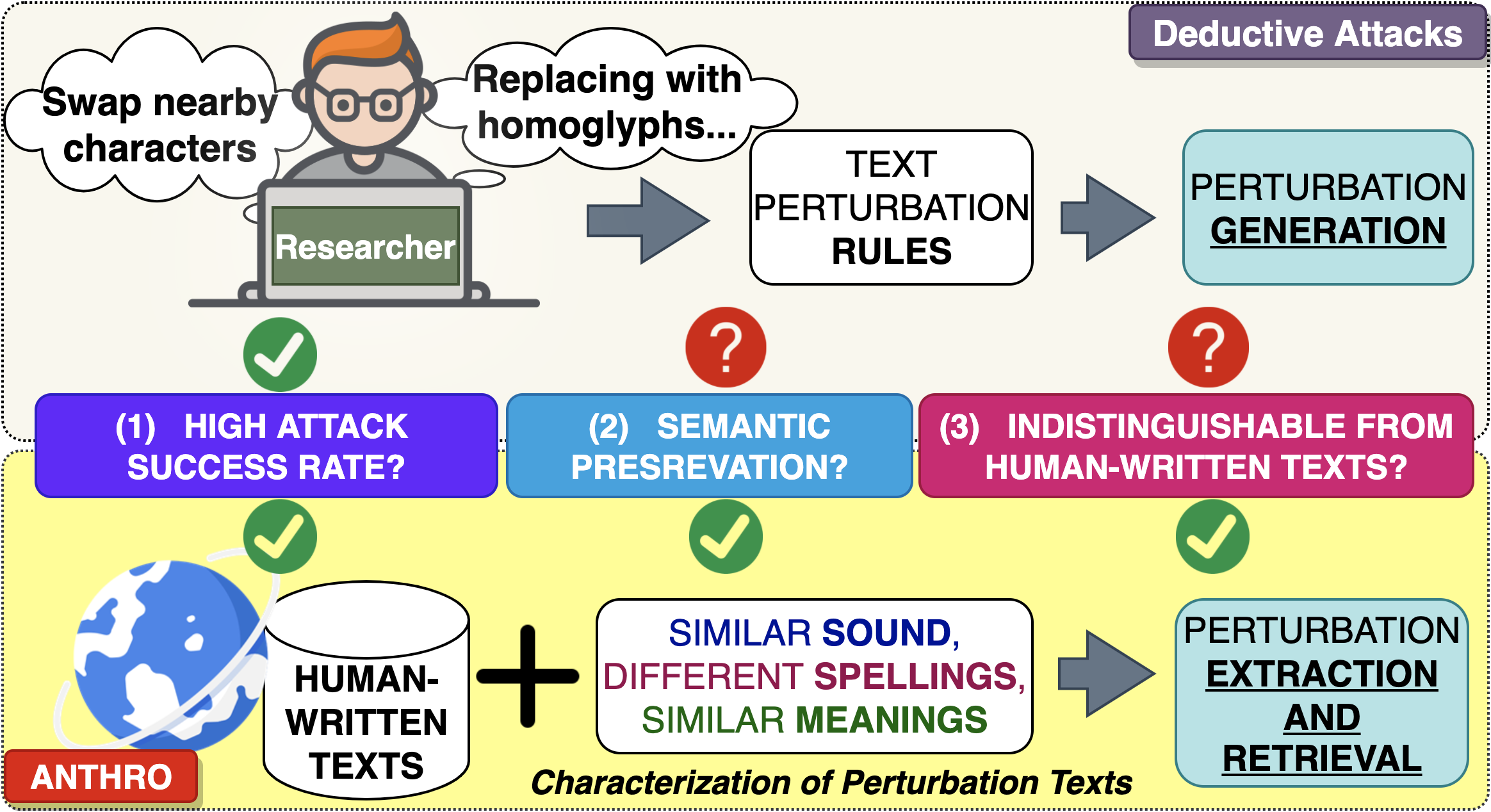}}
\caption{{\mymethod} (Bottom) extracts and uses human-written perturbations for adversarial attacks instead of proposing a specific set of manipulation rules (Top).}
\label{fig:motivation}
% \vspace{-10pt}
\end{figure}

% \renewcommand{\tabcolsep}{1pt}
% \begin{table*}[tb]
%     \centering
%     \footnotesize
%     \begin{tabular}{lccccccc}
%         \toprule
%         \multirow{2}{*}{\textbf{Attacker}} & \textbf{Black} & \textbf{Deductive} & \textbf{Spelling} & \textbf{Inductive} & \textbf{Phonetic} & \textbf{Human} & \multirow{2}{*}{\textbf{Perturbation Example}} \\
%         {} & \textbf{Box} & \textbf{Based} & \textbf{Based} & \textbf{Based} & \textbf{Based} & \textbf{Evaluation} &  \\
%         \cmidrule(lr){1-8}
%         TextBugger~\cite{li2018textbugger} & $\boldcheckmark$ & $\boldcheckmark$ & $\boldcheckmark$ & & & & the dem\textcolor{black}{\uline{\textbf{co}}}rats are very ch\textcolor{black}{\uline{\textbf{ }}}eap\\
%         VIPER~\cite{VIPER} & $\boldcheckmark$ & $\boldcheckmark$ & $\boldcheckmark$ & & & & the d\textcolor{black}{\uline{\textbf{é}}}mocrats \textcolor{black}{\uline{\textbf{å}}}re very ch\textcolor{black}{\uline{\textbf{æ}}}ap\\
%         DeepWordBug~\cite{gao2018black} & $\boldcheckmark$ & $\boldcheckmark$ & $\boldcheckmark$ & & & & the democrats are very cheap\\
%         \cmidrule(lr){1-8}
%         \textbf{{\mymethod}} & $\boldcheckmark$ & $\boldcheckmark$ & $\boldcheckmark$ & $\boldcheckmark$ & $\boldcheckmark$ & $\boldcheckmark$ & the demo\textcolor{black}{\uline{\textbf{kRAT}}}s are very chea\textcolor{black}{\uline{\textbf{a}}}p\\
%         \bottomrule
%     \end{tabular}
%     \caption{Comparison of key features among different character-based adversarial text attacks}
%     \label{ture}
%     \vspace{-10pt}
% \end{table*}
Machine learning (ML) models trained to  optimize only the prediction performance are often vulnerable to adversarial attacks~\cite{papernot2016limitations,wang2019towards}. In the text domain, especially, a character-based adversarial attacker aims to fool a target ML model by generating an adversarial text $x^*$ from an original text $x$ by manipulating characters of different words in $x$, such that some properties of $x$ are preserved~\cite{li2018textbugger,VIPER,gao2018black}. We characterize strong and practical adversarial attacks as three criteria: (1) {\em attack performance}, as measured by the ability to flip a target model's predictions, (2) {\em semantic preservation}, as measured by the ability to preserve the meaning of an original text, and (3) {\em stealthiness}, as measured by how unlikely it is detected as machine-manipulation and removed by defense systems or human examiners (Figure \ref{fig:motivation}). While the first two criteria are natural derivation from adversarial literature~\cite{papernot2016limitations}, stealthiness is also important to be a practical attack under a mass-manipulation scenario. \textcolor{black}{In fact, adversarial text generation remains a challenging task under practical settings.}
%$x^*$ that is perceived as machine-generated and not human-written is more likely to be flagged as suspicious due to the concern of mass-manipulation. 

Previously proposed character-based attacks follow a \textit{deductive} approach where the researchers hypothesize a set of text manipulation strategies that exploit some vulnerabilities of textual ML models (Figure \ref{fig:motivation}). Although these deductively derived techniques can demonstrate superior attack performance, there is no guarantee that they also perform well with regard to semantic preservation and stealthiness. We first analyze why enforcing these properties are challenging especially for character-based attacks.

To preserve the semantic meanings, an attacker can minimize the distance between representative vectors learned from a large pre-trained model--e.g., Universal Sentence Encoder~\cite{cer2018universal} of the two sentences. However, this is only applicable in  word- or sentence-based attacks, not in character-based attacks. It is because character-based manipulated tokens are more prone to become out-of-distribution--e.g., morons$\rightarrow$mor0ns, from what is observed in a typical training corpus where the correct use of English is often assumed. In fact, existing character-based attacks such as \textit{TextBugger}~\cite{li2018textbugger}, \textit{VIPER}~\cite{VIPER} and \textit{DeepWordBug}~\cite{gao2018black} generally assume that the meaning of the original sentence is preserved without further evaluations.

In addition, a robust ML pipeline is often equipped to detect and remove potential adversarial perturbations either via automatic software~\cite{neuspell,pruthi2019combating}, trapdoors~\cite{le2021sweet} or human-in-the-loop~\cite{malcom}. Such detection is feasible especially when the perturbed texts are curated using a set of fixed rules that can be easily re-purposed for defense. Thus, attackers such as \textit{VIPER} and \textit{DeepWordBug}, which map each Latin-based character to either non-English accents (e.g., {\.e}, {\=a}, {\~d}), or homoglyphs (characters of similar shape), fall into this category and can be easily detected %vulnerable 
under simple normalization techniques (Sec. \ref{sec:attack_performance}). \textit{TextBugger}
%~\cite{li2018textbugger} 
circumvents this weakness by utilizing a set of more general character-editing strategies--e.g., replacing and swapping nearby characters to synthesize human-written typos and misspellings. Although texts perturbed by such strategies become less likely to be detected, many of them may distort the meaning of the original text (e.g., ``ga\uline{rb}age"$\rightarrow$``ga\uline{br}age", ``dumb"$\rightarrow$``dub") and can be easily flagged as machine-generated by human examiners. 
Therefore, we argue that
%These examples show that
generating perturbations that both preserve original meanings and are indistinguishable from human-written texts be a critically important yet challenging task. 
% Regardless of any specific manipulation strategies one can propose for adversarial attacks, there is no guarantee if such strategies are also observable from human-written texts. This then undermines ML systems that rely on these attacks for evaluating their robustness against perturbation texts in practice. This even becomes more critical when these systems are trained to detect sensitive domains that are offensive in nature such as hate speech, racism, cyberbullying detection.

%Instead of deriving a set of character manipulation rules to attack,
To overcome these challenges, we introduce \textbf{{\mymethod}}, a novel algorithm that \textit{inductively} finds and extracts text perturbations in the wild. As shown in Figure \ref{fig:motivation}, our method relies on human-written sentences in the Web in their raw form. We then use them to develop a character-based adversarial attack that is not only effective and realistic but is also helpful in training ML models that are more robust against a wide variety of human-written perturbations. Distinguished from previous research, our work considers both spellings and phonetic features (how a word sounds), to characterize text perturbations. Furthermore, we conducted user studies to quantitatively evaluate semantic preservation and stealthiness of adversarial texts. Our contributions are as follows.
\begin{itemize}[leftmargin=\dimexpr\parindent-0.3\labelwidth\relax]
    \setlength\itemsep{-2pt}
    % \item We first show that there is a large distribution gap between manipulation strategies of previous deductive-based adversarial text attacks and human-written perturbations.
    \item {\mymethod} extracts over 600K case-sensitive character-based ``real" perturbations from human-written texts.
    \item {\mymethod} facilitates  black-box adversarial attacks with an average of 82.7\% and 90.7\% attack success rates on BERT and RoBERTa, and drops the \textit{Perspective API}'s precision to only 12\%.
    \item {\mymethod} outperforms the \textit{TextBugger} baseline  by over 50\% in semantic preservation and 40\% in stealthiness in human subject studies.
    %Our human evaluation highlights that {\mymethod} is over 50\% better in terms of semantic preservation, and is also 40\% more likely to be perceived as human-written texts than \textit{TextBugger} baseline. 
    \item {\mymethod} combined with adversarial training also enables BERT classifier to achieve 3\%--14\% improvement in precision over \textit{Perspective API} in understanding human-written perturbations.
\end{itemize}

% \lee{intro section is not very clear--what the main problem is, what's the motivation/justification is. in the page 1, can we think of a figure or illustration that alone explain the problem and/or gist of solution clearly?}

%\section{Preliminary Works and Analysis}
\section{Perturbations in the Wild}
% \renewcommand{\tabcolsep}{1.5pt}
% \begin{table}[t!b]
%     \centering
%     \footnotesize
%     \begin{tabular}{lc}
%         \toprule
%         \textbf{Method} & \textbf{Adversarial Text Example} \\
%         \cmidrule(lr){1-2}
%         VIPER & the \textcolor{black}{\uline{{\~D}{\.e}}}mocr\textcolor{black}{\uline{{\=a}}}ts li\textcolor{black}{\uline{{\.e}}}d to the R\textcolor{black}{\uline{{\"e}}}public\textcolor{black}{\uline{{\.e}}}ns\\
%         DeepWordBug & \\
%         TextBugger & the Demo\textcolor{black}{\uline{rc}}ats lied to the Re\textcolor{black}{\uline{ }} public\textcolor{black}{\uline{@}}ns\\
%         \cmidrule(lr){1-2}
%         \textbf{Human-Written} & the Demo\textcolor{black}{\uline{kRAT}}s lied to the Repub\textcolor{black}{\uline{LIE}}cans \\
%         \bottomrule
%     \end{tabular}
%     \caption{Perturbations of \textit{``the Democrats lied to the Republicans"} by various methods.}
%     \label{tab:examples}
% \end{table}

\subsection{Machine v.s. Human Perturbations} \label{sec:preliminary_analysis}

Perturbations that are neither natural-looking nor resembling human-written texts are more likely to be detected by defense systems (thus not a practical attack from adversaries' perspective).
%trigger a robust ML system to flag them as suspicious and hence remove them. 
%This is applicable especially in critical domains such as disinformation or toxicity detection where mass machine-manipulation can be very detrimental. 
However, some existing character-based perturbation strategies, including \textit{TextBugger}, \textit{VIPER} and \textit{DeepWordBug}, follow a \textit{deductive} approach and their generated texts often do not resemble human-written texts. 
% In fact, our human-study (Sec. \ref{sec:human_study}) later shows that human examiners can often differentiate a machine-generated typo--e.g., ``stupid"$\rightarrow$``sutpid", from a genuine human-written one--e.g., ``stupid"$\rightarrow$``stuupid". 
% If their perturbations successfully resemble human-written texts, it becomes more difficult for real-life ML systems to detect and remove them. However, we observe that such tactics are very different from those employed by humans.
% This section then analyzes the discrepancy between machine-generated and human-written texts.
% However, this deductive mechanism does not guarantee that the generated perturbations are observable in real-life texts--i.e., ensemble human-written texts.
Qualitatively, however, we find that humans express much more diverse and creative~\cite{tagg2011wot} perturbations (Figure \ref{fig:wordcloud}, Appendix) than ones generated by such deductive approaches. For example, humans frequently (1) capitalize and change the parts of a word to emphasize distorted meanings (e.g.,``democrats``$\rightarrow$``democRATs", ``republicans"$\rightarrow$``republiCUNTs"), (2) hyphenate a word (e.g., ``depression"$\rightarrow$``de-pres-sion"), (3) use emoticons to emphasize meaning (e.g., ``shit"$\rightarrow$``sh$\poopemoji$t"), (4) repeat particular characters (e.g., ``dirty"$\rightarrow$``diiirty", ``porn"$\rightarrow$``pooorn"), or (5) insert phonetically similar characters (e.g., ``nigger"$\rightarrow$``nighger"). \textcolor{black}{Human-written perturbations do not manifest any fixed rules and often require some context understanding. Moreover, one can generate a new meaningful perturbation simply by repeating a character--e.g., ``porn"$\rightarrow$``pooorn". Thus, it is challenging to systematically generate all such perturbations, if not impossible. Moreover, it is very difficult for spell-checkers, which usually rely on a fixed set of common spelling mistakes and an edit-distance threshold, to correct and detect all human-written perturbations.}

We later show that human examiners rely on personal exposure from Reddit or YouTube comments to decide if a word choice looks natural (Sec. \ref{sec:human_study}). Quantitatively, we discover that not all the perturbations generated by deductive methods are observed on the Web (Table \ref{tab:analysis_reallife}). To analyze this, we first use each attack to generate all possible perturbations of either (1) a list of over 3K unique offensive words or (2) a set of the top 5 offensive words (``c*nt'', ``b*tch'', ``m*therf***er'', ``bast*rd'', ``d*ck''). Then, we calculate how many of the perturbed words are present in a dataset of over 34M online news comments or are used by at least 50 unique commentators on Reddit, respectively. Even though \textit{TextBugger} was well-known to simulate human-written typos as adversarial texts, merely 51.6\% and 7.1\% of its perturbations are observed on Reddit and online news comments, implying \textit{TextBugger}'s generated adversarial texts being ``unnatural" and ``easily-detectable" by human-in-the-loop defense systems.
%Moreover, using homoglyphs and non-English accents is not intuitive among human writers.

\renewcommand{\tabcolsep}{2.5pt}
\begin{table}[tb]
    \centering
    \footnotesize
    \begin{tabular}{lcc}
         \toprule
         \textbf{Attacker} & \textbf{Reddit Comts.} & \textbf{News Comts.} \\
         \textbf{\#texts, \#tokens} & \textbf{>>5B, N/A} & \textbf{(34M, 11M)} \\
         \cmidrule(lr){1-3}
        TextBugger & \uline{51.6\%} (126/244) & \uline{7.10\%} (11K/152K)\\
        VIPER & 3.2\% (1/31) & 0.13\% (25/19K) \\
        DeepWordBug & 0\% (0/31) & 0.27\% (51/19K)\\
        \cmidrule(lr){1-3}
        ANTHRO & \textbf{82.4\%} (266/323) & \textbf{55.7\%} (16K/29K)\\
         \bottomrule
    \end{tabular}
    \caption{Percentage of offensive perturbed words generated by different attacks that can be observed in real human-written comments on Reddit and online news.
    % \lee{If not observed, it's more likely to be detected? can the opposite (ie, perturbation not often observed is more easily detectable?) be true too? any backing literature?}
    }
    \label{tab:analysis_reallife}
    % \vspace{-10pt}
\end{table}

%That being said, we propose {\mymethod} algorithm, which directly utilizes human-written perturbations for adversarial attack. These perturbations are highly overlapped with \textit{unseen} human-written texts on Reddit and online news (Table \ref{tab:analysis_reallife}).

% Different from the deductive approaches, inductive method first characterizes ``what text perturbations are" from observations, then use such characterization to search for all perturbations directly from human-written texts. This will enable us to curate perturbations that better resemble human behaviors. We further propose {\mymethod} algorithm, which combines both deductive and inductive approach to expand the search space. Table \ref{tab:analysis_reallife} shows that {\mymethod} can generate perturbed words that are highly overlapped with \textit{unseen} human written-texts on Reddit and online news.

\subsection{The SMS Property: Similar Sound, Similar Meaning, Different Spelling} 

The existence of a non-arbitrary relationship between sounds and meanings has been proven by a life-long research establishment~\cite{kohler1967gestalt,jared1991does,gough1972one}. In fact, \citet{blasi2016sound} analyzed over 6K languages and discovered a high correlation between a word's sound and meaning both inter- and intra-cultures. \citet{aryani2020affective} found that how a word sounds links to an individual's emotion. This motivates us to hypothesize that words spelled differently yet have the same meanings such as text perturbations will also have similar sounds.

Figure \ref{fig:wordcloud} (Appendix) displays several perturbations that are found from real-life texts. Even though these perturbations are \textit{spelled differently} from the original word, they all preserve \textit{similar meanings} when perceived by humans. Such semantic preservation is feasible because humans perceive these variations \textit{phonetically similar} to the respective original words~\cite{van1987rows}. For example, both ``republican" and ``republi\uline{k}an" sound similar when read by humans. Therefore, given the surrounding context of a perturbed sentence--e.g., ``\textit{President Trump is a} republi\uline{k}an'', and the phonetic similarity of ``republican'' and ``republi\uline{k}an'', end-users are more likely to interpret the perturbed sentence as ``\textit{President Trump is a republican}''. We call these characteristics of text perturbations the \textit{SMS} property: ``\textit{similar \uline{S}ound, similar \uline{M}eaning, different \uline{S}pellings}''. Noticeably, the SMS characterization includes a subset of ``visually similar" property of perturbations as studied in previous adversarial attacks such as \textit{TextBugger} (e.g., ``hello'' sounds similar with ``he\uline{11}o''), \textit{VIPER} and \textit{DeepWordBug}. However, two words that look very similar sometimes carry different meanings--e.g., ``garbage''$\rightarrow$``gabrage''. Moreover, our characterization is also distinguished from \textit{homophones} (e.g., ``to'' and ``two'') which describe words with similar sound yet \textit{different meaning}.

\section{A Realistic Adversarial Attack}
Given the above analysis, we now derive our proposed {\mymethod} adversarial attack. We first share how to systematically encode the sound--i.e., phonetic feature, of any given words and use it to search for their human-written perturbations that satisfy the SMS property. Then, we introduce an iterative algorithm that utilizes the extracted perturbations to attack textual ML models.

\subsection{Mining Perturbations in the Wild}\label{sec:mining_perturbations}

\noindent \textbf{Sound Encoding with \textsc{Soundex++.}} To capture the sound of a word, we adopt and extend the case-insensitive \textsc{Soundex} algorithm. \textsc{Soundex} helps index a word based on how it sounds rather than how it is spelled~\cite{stephenson1980methodology}. Given a word, \textsc{Soundex} first keeps the 1st character. Then, it removes all vowels and matches the remaining characters \textit{one by one} to a digit following a set of predefined rules--e.g., ``B'', ``F''$\rightarrow$1, ``D'', ``T''$\rightarrow$3~\cite{stephenson1980methodology}. For example, ``Smith'' and ``Smyth'' are both encoded as \textsc{S530}.

As the \textsc{Soundex} system was designed mainly for encoding surnames, it does not necessarily work for texts in the wild. For example, it cannot recognize visually-similar perturbations such as ``l"$\rightarrow$``1", ``a"$\rightarrow$``@" and ``O"$\rightarrow$``0". Moreover, it always fixes the 1st character as part of the final encodes. This rule is too rigid and can result in words that are entirely different yet encoded the same (Table \ref{tab:soundexplus}). To solve these issues, we propose a new \textsc{Soundex++}
% \lee{Soundex++ is better like C++ is next to C without C+}
algorithm. \textcolor{black}{\textsc{Soundex++} is equipped to both recognize visually-similar characters and encode the sound of a word at different hierarchical levels $\mathbf{k}$ (Table \ref{tab:soundexplus})}. Particularly, at level $\mathbf{k}{=}0$, \textsc{Soundex++} works similar to \textsc{Soundex} by fixing the first character. At level $\mathbf{k}{\geq}1$, \textsc{Soundex++} instead fixes the first $\mathbf{k}{+}1$ characters and encodes the rest.
\vspace{5pt}

\renewcommand{\tabcolsep}{4pt}
\begin{table}[tb]
    \centering
    \footnotesize
    \begin{tabular}{lcc}
        \toprule
         \textbf{Word} & \multicolumn{1}{c}{\textbf{\textsc{Soundex}}} & \multicolumn{1}{c}{\textbf{\textsc{Soundex++} (Ours)}} \\
         \cmidrule(lr){1-3}
         porn & P650 & P650 ($\mathbf{k}{=}0$), PO650 ($\mathbf{k}{=}1$)\\
        %  po\textcolor{black}{\uline{or}}rn & P650 & (same as above) \\
         p\textcolor{black}{\uline{0}}rn & \uline{P065}(\xmark) & (same as above) \\
         \cmidrule(lr){1-3}
         lesbian & L215 & L245 ($\mathbf{k}{=}0$), LE245 ($\mathbf{k}{=}1$) \\
         lesb\textcolor{black}{\uline{b}}i\textcolor{black}{\uline{@}}n & \uline{L21@}(\xmark) & (same as above) \\
         l\textcolor{red}{\uline{o}}sbian & L215(\xmark) & L245 ($\mathbf{k}{=}0$), \uline{LO245} ($\mathbf{k}{=}1$) \\
        \bottomrule
        \multicolumn{3}{l}{(\xmark): Incorrect encoding}
    \end{tabular}
    \caption{\textsc{Soundex++} can capture visually similar characters and is more accurate in differentiating between desired (\textcolor{black}{blue}) and undesired (\textcolor{red}{red}) perturbations.
% \lee{cound't fully follow  this table--unclear to me??}
    }
    \label{tab:soundexplus}
    % \vspace{-5pt}
\end{table}

\begin{table}[tb]
    \centering
    \footnotesize
    {\color{black}
    \begin{tabular}{cccccc}
        \toprule
        \textbf{Key} & TH000 & DE5263 & AR000 & DI630 & NO300\\
        \cmidrule(lr){1-6}
        \textbf{Value} & the & democrats & are & dirty & not\\
        \textbf{(Set)} & & demokRATs & arre & dirrrty\\
        \cmidrule(lr){1-6}
        \multicolumn{6}{l}{\textcolor{red}{\mymethod}(democrats,$\mathbf{k}{=}1$,$\mathbf{d}{=}1$)$\rightarrow$\{democrats, demokRATs\}} \\
        \multicolumn{6}{l}{\textcolor{red}{\mymethod}(dirty,$\mathbf{k}{=}1$,$\mathbf{d}{=}2$)$\rightarrow$\{dirty, dirrrty\}} \\
        \bottomrule
    \end{tabular}
    }
    \caption{\textcolor{black}{Examples of hash table $H_1(k{=}1)$ curated from sentences \textit{``the demokRATs are dirrrty"} and \textit{``the democrats arre not dirty"} and its utilization.}}
    \label{tab:mining_example}
    % \vspace{-15pt}
\end{table}

\begin{figure}[tb]
% https://app.diagrams.net/#G1l9fJ6BC3A8MPywE2m6hxkYrXsVI4AqUr
\centerline{\includegraphics[width=0.45\textwidth]{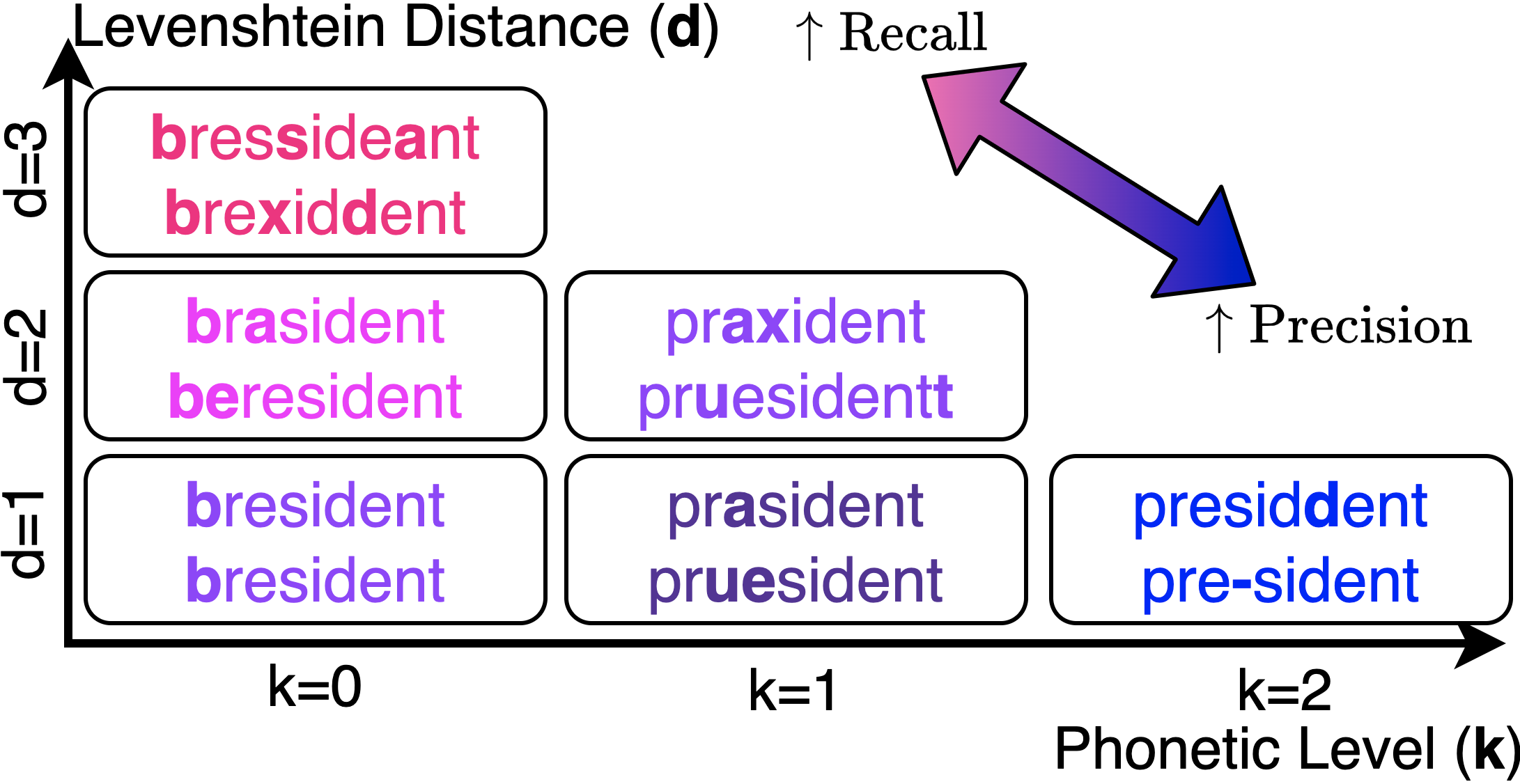}}
\caption{Trade-off between precision and recall of extracted perturbations for the word ``president" w.r.t different $\mathbf{k}$ and $\mathbf{d}$ values. Higher $\mathbf{k}$ and lower $\mathbf{d}$ associate with better preservation of the original meaning.}
\label{fig:precision_recall}
% \vspace{-10pt}
\end{figure}

\noindent \textbf{Levenshtein Distance $\mathbf{d}$ and Phonetic Level $\mathbf{k}$ as a Semantic Preservation Proxy.} Since \textsc{Soundex++} is not designed to capture a word's semantic meaning, we utilize both phonetic parameter $\mathbf{k}$ and \textit{Levenshtein distance} $\mathbf{d}$ ~\cite{levenshtein1966binary} as a heuristic approximation to measure the semantic preservation between two words. Intuitively, the higher the phonetic level ($\mathbf{k}{\geq}1$) at which two words share the same \textsc{Soundex++} code and the smaller the Levenshtein distance $\mathbf{d}$ to transform one word to another, the more likely human associate them with the meaning. In other words, $\mathbf{k}$ and $\mathbf{d}$ are hyper-parameters that help control the trade-off between precision and recall when retrieving perturbations of a given word such that they satisfy the SMS property (Figure \ref{fig:precision_recall}). We will later carry out a human study to evaluate how well our extracted perturbations can preserve the semantic meanings in practice.
\vspace{5pt}

\noindent \textbf{Mining from the Wild.} To mine all human-written perturbations, we first collect a large corpus $\mathcal{D}$ of over 18M sentences written by netizens from 9 different datasets (Table \ref{tab:dataset} in Appendix). We select these datasets because they include offensive texts such as hate speech, sensitive search queries, etc., and hence very likely to include text perturbations. Next, for each phonetic level $\mathbf{k}{\leq}K$, we curate different hash tables $\{H\}_0^K$ that maps a unique \textsc{Soundex++} code $\mathbf{c}$ to a set of its matching unique \textit{case-sensitive} tokens that share the same encoding $\mathbf{c}$ as follows:
\begin{equation}
\begin{aligned}
    H_\mathbf{k}: \mathbf{c} \mapsto \{&w_j | S(w_i,k)=S(w_j,k)=\mathbf{c}  \\
    &\bigforall w_i, w_j \in \mathcal{D}, w_i \neq w_j\},
\end{aligned}
\label{eqn:extraction}
\end{equation}
\noindent where $S(w,\mathbf{k})$ returns the \textsc{Soundex++} code of token $w$ at phonetic level $\mathbf{k}$, $K$ is the largest phonetic level we want to encode. With $\{H\}_0^K$, $\mathbf{k}$ and $\mathbf{d}$, we can now search for the set of perturbations $G_\mathbf{k}^\mathbf{d}(w^*)$ of a specific target token $w^*$ as follows:
\begin{equation}
\begin{aligned}
    G_\mathbf{k}^\mathbf{d}(w^*){\leftarrow}\{&w_j | w_j{\in}H_\mathbf{k}[S(w^*,k)],\\
    &\mathrm{Lev}(w^*,w_j){\leq}\mathbf{d}\}
    \label{eqn:searching}
\end{aligned}
\end{equation}
\noindent where $\mathrm{Lev}(w^*,w_j)$ returns the Levenshtein distance between $w^*$ and $w^j$. Noticeably, we only extract $\{H\}_0^K$ \textbf{once} from $\mathcal{D}$ via Eq. (\ref{eqn:extraction}), then we can use Eq. (\ref{eqn:searching}) to retrieve all perturbations for a given word during deployment. We name this method of mining and retrieving human-written text perturbations in the wild as \textbf{\mymethod}, aka \textit{human-like} perturbations:
\begin{equation}
    \textcolor{red}{\mathrm{\mymethod}}: w*,\mathbf{k},\mathbf{d}, \{H\}_0^K \;\; \longmapsto\;\; G_\mathbf{k}^\mathbf{d}(w^*)
    \label{eqn:anthro}
\end{equation}

\begin{algorithm}[t!b]
 \caption{{\mymethod} Attack Algorithm}
%  \lee{Score() function undefined, or is this S() function in Eq 1?}}
 \label{alg:attack}
 \begin{algorithmic}[1]
\STATE \textbf{Input:} $\{H\}_0^K$, $\mathbf{k}$, $\mathbf{d}$ \\
\STATE \textbf{Input:} target classifier $f$, original sentence $x$ \\
\STATE \textbf{Output:} perturbed sentence $x^*$ \\
\STATE \textit{Initialize:} $x^*\leftarrow x$\\
% \STATE \textcolor{black}{// Calculate importance score and rank words}
\STATE \textbf{for} word $x_i$ \textbf{in} $x$ \textbf{do}:$\quad s_i{\leftarrow}\mathrm{Score}(x_i, f)$ \\
\STATE $\mathcal{W}_\mathrm{order}{\leftarrow}\mathrm{Sort}(x_1, x_2,..x_m)$ according to $s_i$ \\
% \STATE \textcolor{black}{// Replace the best perturbations for each word}\\
\STATE \textbf{for} $x_i$ in $\mathcal{W}_\mathrm{order}$ \textbf{do}:
\STATE $\quad \mathcal{P}{\leftarrow}\mathrm{\textcolor{red}{{\mymethod}}}(x_i,\mathbf{k},\mathbf{d},\{H\}_0^K)$ \textcolor{black}{// Eq.(\ref{eqn:anthro})}
\STATE $\quad x^*{\leftarrow}$ replace $x_i \in x$ with the best $w \in \mathcal{P}$
\STATE $\quad $\textbf{if} $f(x^*){\neq}f(x)$ \textbf{then} return $x^*$
\STATE \textbf{return} None
 \end{algorithmic}
%  \vspace{-10pt}
\end{algorithm}

\noindent \textbf{{\mymethod} Attack.} To utilize {\mymethod} for adversarial attack on model $f(x)$, we propose the {\mymethod} attack algorithm (Alg. \ref{alg:attack}). We use the same iterative mechanism (Ln.9--13) that is common among other black-box attacks. This process replaces the most vulnerable word in sentence $x$, which is evaluated with the support of $\mathrm{Score)(\cdot)}$ function (Ln. 5), with the perturbation that best drops the prediction probability $f(x)$ on the correct label. Unlike the other methods, {\mymethod} inclusively draws from perturbations extracted from human-written texts captured in $\{\mathcal{H}\}_0^K$ (Ln. 10). We adopt the $\mathrm{Score}(\cdot)$ from \textit{TextBugger}.

\section{Evaluation}
We evaluate {\mymethod} by:
%This section evaluates and compares {\mymethod} with other character-based attack methods on three aspects, namely 
(1) attack performance, (2) semantic preservation, and (3) human-likeness--i.e., how likely an attack message is spotted as machine-generated by human examiners. 
%not to be detected as machine and not human-written.
% \lee{the term Adaku used was "reverse TT"} 
% \lee{in intro, you mentioned 3 reasons which are slightly different from these 3 aspects in evaluation--especially, how "stealthiness" can be captured by RTT?}
% Specifically, we evaluate how effective they are under different defense mechanisms, how well they can preserve the original meanings, and if they are indistinguishable from human-written texts.

\subsection{Attack Performance} \label{sec:attack_performance}
\noindent \textbf{Setup.} We use BERT (\textit{case-insensitive})~\cite{jin2019bert} and RoBERTa (\textit{case-sensitive})~\cite{liu2019roberta} as target classifiers to attack. We evaluate on three public tasks, namely detecting toxic comments ((TC) dataset, Kaggle 2018), hate speech ((HS) dataset~\cite{hateoffensive}), and online cyberbullying texts ((CB) dataset~\cite{cyberbullyingdata}). We split each dataset to \textit{train}, \textit{validation} and \textit{test} set with the 8:1:1 ratio. Then, we use the train set to fine-tune BERT and RoBERTa with a maximum of 3 epochs and select the best checkpoint using the validation set. BERT and RoBERTa achieve around 0.85--0.97 in F1 score on the test sets (Table \ref{tab:attack_dataset} in Appendix). We evaluate with targeted attack (change positive$\rightarrow$negative label) since it is more practical. We randomly sample 200 examples from each test set and use them as initial sentences to attack. We repeat the process 3 times with unique random seeds and report the results. We use the \textit{attack success rate (Atk\%)} metric--i.e., the number of examples whose labels are flipped by an attacker over the total number of texts that are correctly predicted pre-attack. We use the 3rd party open-source \textit{OpenAttack}~\cite{zeng2020openattack} framework to run all evaluations.
\vspace{5pt}

\noindent \textbf{Baselines.} We compare {\mymethod} with three baselines, namely \textit{TextBugger}~\cite{li2018textbugger}, \textit{VIPER}~\cite{VIPER} and \textit{DeepWordBug}~\cite{gao2018black}. These attackers utilize different character-based manipulations to craft their adversarial texts as described in Sec. \ref{sec:introduction}. From the analysis in Sec. \ref{sec:mining_perturbations} and Figure \ref{fig:precision_recall}, we set $\mathbf{k}{\leftarrow}1$ and $\mathbf{d}{\leftarrow}1$ for {\mymethod} to achieve a balanced trade-off between precision and recall on the SMS property. We examine all attackers under several combinations of different normalization layers. They are (1) \textit{\uline{A}ccents normalization} (A) and (2) \textit{\uline{H}omoglyph normalization}~\footnote{\underline{https://github.com/codebox/homoglyph}} (H), which converts non-English accents and homoglyphs to their corresponding ascii characters, (3) \textit{\uline{P}erturbation normalization} (P), which normalizes potential character-based perturbations using the SOTA misspelling correction model \textit{Neuspell}~\cite{neuspell}. These normalizers are selected as counteracts against the perturbation strategies employed by \textit{VIPER} (uses non-English accents), \textit{DeepWordBug} (uses homoglyphs) and \textit{TextBugger}, {\mymethod} (based on misspelling and typos), respectively.
\vspace{5pt}

\renewcommand{\tabcolsep}{7pt}
\begin{table*}[tb]
    \centering
    \footnotesize
    \begin{tabular}{lccccccc}
    \toprule
         \multirow{3}{*}{\textbf{\textbf{Attacker}}} & \multirow{3}{*}{\textbf{Normalizer}} & \multicolumn{3}{c}{\textbf{BERT} (\textit{case-insensitive})} & \multicolumn{3}{c}{\textbf{RoBERTa} (\textit{case-sensitive)}} \\
         \cmidrule(lr){3-5}\cmidrule(lr){6-8}
        {} & {} & \textbf{TC} & \textbf{HS} & \textbf{CB} & \textbf{TC} & \textbf{HS} & \textbf{CB} \\
        % \cmidrule(lr){1-1}\cmidrule(lr){2-2}\cmidrule(lr){3-3}\cmidrule(lr){4-4}\cmidrule(lr){5-5}\cmidrule(lr){6-6}\cmidrule(lr){7-7}\cmidrule(lr){8-8}
        % {} & \multicolumn{6}{c}{---Default Settings---} \\
        \cmidrule(lr){1-8}
        TextBugger & - & \textbf{0.76$\pm$0.02} &\textbf{0.94$\pm$0.01} &\textbf{0.78$\pm$0.03} & 0.77$\pm$0.06 & 0.87$\pm$0.01 & 0.72$\pm$0.01 \\
        DeepWordBug & - & 0.56$\pm$0.04 &0.68$\pm$0.01 &0.50$\pm$0.02 & 0.52$\pm$0.01 & 0.42$\pm$0.04 & 0.38$\pm$0.04 \\
        VIPER & - & \textcolor{red}{0.08}$\pm$\textcolor{red}{0.03} & \textcolor{red}{0.01}$\pm$\textcolor{red}{0.01} &\textcolor{red}{0.13}$\pm$\textcolor{red}{0.02} & \textbf{1.00$\pm$0.00} & \textbf{1.00$\pm$0.00} & \textbf{0.99$\pm$0.01} \\
        \textbf{{\mymethod}} & - & \uline{0.72$\pm$0.02} & \uline{0.82$\pm$0.01} & \uline{0.71$\pm$0.02} & \uline{0.84$\pm$0.00} & \uline{0.93$\pm$0.01} & \uline{0.78$\pm$0.01} \\
        % \textbf{{\mymethodx}} & \textbf{0.82$\pm$0.01} & \textbf{0.97$\pm$0.01} & \textbf{0.88$\pm$0.04} & \uline{0.91$\pm$0.02} & \uline{0.97$\pm$0.01} & \uline{0.89$\pm$0.02} \\
        
        \cmidrule(lr){1-8}
        % {} & \multicolumn{6}{c}{---With \uline{Accents Normalization} (during training and inference)---} \\
        TextBugger  & A & - & - & - & \uline{0.72$\pm$0.02} & \uline{0.92$\pm$0.00} & \uline{0.74$\pm$0.02}\\
        DeepWordBug & A & - & - & - & 0.43$\pm$0.02 & 0.59$\pm$0.03 & 0.43$\pm$0.01 \\
        VIPER & A & - & - & - & \textcolor{red}{0.09}$\pm$\textcolor{red}{0.01} & \textcolor{red}{0.05}$\pm$\textcolor{red}{0.01} & 0.17$\pm$0.02 \\
        % \cmidrule(lr){1-7}
        \textbf{{\mymethod}} & A & - & - & - & \textbf{0.77$\pm$0.02} & \textbf{0.94$\pm$0.02} & \textbf{0.84$\pm$0.02} \\
        % \textbf{{\mymethodx}} & - & - & - & \textbf{0.84$\pm$0.02} & \textbf{0.98$\pm$0.00} & \textbf{0.91$\pm$0.02} \\
        
        \cmidrule(lr){1-8}
        % {} & \multicolumn{6}{c}{---With \uline{Homoglyph + Accents Normalization} (during training and inference)---} \\
        % {} & \multicolumn{6}{c}{---(during training and inference)---} \\
        TextBugger & A+H & \textbf{0.78$\pm$0.03} &\textbf{0.85$\pm$0.00} &\textbf{0.79$\pm$0.00} & \uline{0.74$\pm$0.02} & \uline{0.93$\pm$0.01} & \uline{0.77$\pm$0.03} \\
        DeepWordBug & A+H & \textcolor{red}{0.04}$\pm$\textcolor{red}{0.00} &\textcolor{red}{0.06}$\pm$\textcolor{red}{0.02} &\textcolor{red}{0.01}$\pm$\textcolor{red}{0.01} & \textcolor{red}{0.03}$\pm$\textcolor{red}{0.01} & \textcolor{red}{0.01}$\pm$\textcolor{red}{0.01} & \textcolor{red}{0.06}$\pm$\textcolor{red}{0.02}\\
        VIPER & A+H & \textcolor{red}{0.07}$\pm$\textcolor{red}{0.00} & \textcolor{red}{0.01}$\pm$\textcolor{red}{0.01} &\textcolor{red}{0.10}$\pm$\textcolor{red}{0.00} & \textcolor{red}{0.13}$\pm$\textcolor{red}{0.02} & \textcolor{red}{0.07}$\pm$\textcolor{red}{0.01} & 0.17$\pm$0.01 \\
        % \cmidrule(lr){1-7}
        \textbf{{\mymethod}} & A+H & \uline{0.76$\pm$0.02} &\uline{0.77$\pm$0.03} &\uline{0.73$\pm$0.05} & \textbf{0.82$\pm$0.02} & \textbf{0.97$\pm$0.00} & \textbf{0.82$\pm$0.02}\\
        % \textbf{{\mymethodx}} & \textbf{0.84}$\pm$\textbf{0.03} &\textbf{0.91}$\pm$\textbf{0.02} &\textbf{0.88}$\pm$\textbf{0.01} & \textbf{0.87}$\pm$\textbf{0.03} & \textbf{0.99}$\pm$\textbf{0.00} & \textbf{0.90}$\pm$\textbf{0.02} \\
        
        \cmidrule(lr){1-8}
        % {} & \multicolumn{6}{c}{---With \uline{Misspellings Correction + Homoglyph + Accents Normalization}---} \\
        % {} & \multicolumn{6}{c}{---(during training and inference)---} \\
        TextBugger & A+H+P & \textbf{0.73$\pm$0.02} & \textbf{0.64$\pm$0.06} & \textbf{0.70$\pm$0.04} &0.68$\pm$0.06 &0.57$\pm$0.03 &0.66$\pm$0.04 \\
        DeepWordBug & A+H+P & \textcolor{red}{0.02$\pm$0.01} & \textcolor{red}{0.04$\pm$0.02} & \textcolor{red}{0.01$\pm$0.01} &\textcolor{red}{0.02$\pm$0.01} &\textcolor{red}{0.01$\pm$0.01} &\textcolor{red}{0.02$\pm$0.01} \\
        VIPER & A+H+P & \textcolor{red}{0.12$\pm$0.01} & \textcolor{red}{0.04$\pm$0.01} & 0.17$\pm$0.03 &\textcolor{red}{0.11$\pm$0.02} &\textcolor{red}{0.05$\pm$0.01} &0.18$\pm$0.01 \\
        % \cmidrule(lr){1-7}
        \textbf{{\mymethod}} & A+H+P & \uline{0.65$\pm$0.04} & \textbf{0.64$\pm$0.01} & \uline{0.60$\pm$0.05} &\textbf{0.80$\pm$0.02} &\textbf{0.91$\pm$0.03} &\textbf{0.82$\pm$0.02} \\
        % \textbf{{\mymethodx}} & \textbf{0.85$\pm$0.04} & \textbf{0.79$\pm$0.02} & \textbf{0.84$\pm$0.03} &\textbf{0.88$\pm$0.04} &\textbf{0.93$\pm$0.01} &\textbf{0.91$\pm$0.01} \\
        
    \bottomrule
    \multicolumn{8}{l}{(-) BERT already has the accents normalization (A normalizer) by default, \textcolor{red}{(Red)}: Poor performance (Atk\%<0.15)}
    \end{tabular}
    \caption{Averaged attack success rate (Atk\%$\uparrow$) of different attack methods}
    \label{tab:results}
    % \vspace{-10pt}
\end{table*}

\noindent \textbf{Results.} Overall, both {\mymethod} and \textit{TextBugger} perform the best. \textcolor{black}{Being case-sensitive, {\mymethod} performs significantly better on RoBERTa and is competitive on BERT when compared to \textit{TextBugger} (Table \ref{tab:results}). This happens because RoBERTa is case-sensitive (unlike the \textit{base-uncased-bert} BERT model we used) and only {\mymethod} is case-sensitive out of all attack baselines. For example, the perturbation ``democrats"$\rightarrow$``democRATs" is considered as a perturbation for RoBERTa but not for other case-insensitive models. This gives {\mymethod} an advantage in practice because many popular commercial API services (e.g., the popular \textit{Perspective API}, the sentiment analysis and text categorization API from Google) are case-sensitive--i.e., ``democrats"$\neq$``democRATs". (See more at Table \ref{tab:non_abusive}).}

\textit{VIPER} achieves a near perfect score on RoBERTa, yet it is ineffective on BERT because RoBERTa uses the accent \.G as a part of its byte-level BPE encoding~\cite{liu2019roberta} while BERT by default removes all such accents. Since \textit{VIPER} exclusively utilizes accents, its attacks can be easily corrected by the \textit{accents normalizer} (Table \ref{tab:results}). Similarly, \textit{DeepWordBug} perturbs texts with homoglyph characters, most of which can also be normalized using a 3rd party homoglyph detector (Table \ref{tab:results}).

In contrast, even under all normalizers--i.e., A+H+P, \textit{TextBugger} and {\mymethod} still achieves 66.3\% and 73.7\% in Atk\% on average across all evaluations. Although \textit{Neuspell}~\cite{neuspell} drops \textit{TextBugger}'s Atk\% 14.7\% across all runs, it can only reduce the Atk\% of {\mymethod} a mere 7.5\% on average. This is because \textit{TextBugger} and \textit{Neuspell} or other dictionary-based typo correctors rely on fixed deductive rules--e.g., swapped, replaced by neighbor letters, for attack and defense. However, {\mymethod} utilizes human-written perturbations which are greatly varied, hence less likely to be systematically detected. \textcolor{black}{We further discuss the limitation of misspelling correctors such as NeuSpell in Sec. \ref{sec:discussion}.}
% Due to its case-sensitive attack, {\mymethod} also consistently outperforms \textit{TextBugger} on RoBERTa. 

\renewcommand{\tabcolsep}{1.5pt}
\begin{table*}[tb]
    \centering
    \footnotesize
    \begin{tabular}{lccccccc}
    \toprule
         \multirow{3}{*}{\textbf{\textbf{Attacker}}} & \multirow{3}{*}{\textbf{Normalizer}} & \multicolumn{3}{c}{\textbf{BERT} (\textit{case-insensitive})} & \multicolumn{3}{c}{\textbf{RoBERTa} (\textit{case-sensitive)}} \\
         \cmidrule(lr){3-5}\cmidrule(lr){6-8}
        {} & {} & \textbf{Toxic Comments} & \textbf{HateSpeech} & \textbf{Cyberbullying} & \textbf{Toxic Comments} & \textbf{HateSpeech} & \textbf{Cyberbullying} \\
        \cmidrule(lr){1-8}
        TextBugger & - & \uline{0.76$\pm$0.02} &\uline{0.94$\pm$0.01} &\uline{0.78$\pm$0.03} & 0.77$\pm$0.06 & 0.87$\pm$0.01 & 0.72$\pm$0.01 \\
        % \textbf{{\mymethod}} & - & 0.72$\pm$0.02 &0.82$\pm$0.01 &0.71$\pm$0.02 & {0.84}$\pm${0.00} & {0.93}$\pm${0.01} & {0.78}$\pm${0.01} \\
        \textbf{{\mymethodx}} & - & \textbf{0.82$\pm$0.01} & \textbf{0.97$\pm$0.01} & \textbf{0.88$\pm$0.04} & \textbf{0.91$\pm$0.02} & \textbf{0.97$\pm$0.01} & \textbf{0.89$\pm$0.02} \\
        \cmidrule(lr){1-8}
        TextBugger & A+H+P & \uline{0.73$\pm$0.02} & \uline{0.64$\pm$0.06} & \uline{0.70$\pm$0.04} &0.68$\pm$0.06 &0.57$\pm$0.03 &0.66$\pm$0.04 \\
        % \textbf{{\mymethod}} & A+H+P & 0.65$\pm$0.04 & \uline{0.64$\pm$0.01} & 0.60$\pm$0.05 &\uline{0.80$\pm$0.02} &\uline{0.91$\pm$0.03} &\uline{0.82$\pm$0.02} \\
        \textbf{{\mymethodx}} & A+H+P & \textbf{0.85$\pm$0.04} & \textbf{0.79$\pm$0.02} & \textbf{0.84$\pm$0.03} &\textbf{0.88$\pm$0.04} &\textbf{0.93$\pm$0.01} &\textbf{0.91$\pm$0.01} \\
    \bottomrule
    \end{tabular}
    \caption{Averaged attack success rate (Atk\%$\uparrow$) of {\mymethodx} and \textit{TextBugger}}
    \label{tab:results2}
    % \vspace{-10pt}
\end{table*}

\subsection{Human Evaluation}\label{sec:human_study}
Since {\mymethod} and \textit{TextBugger} are the top two effective attacks, this section will focus on evaluating their ability in semantic preservation and human-likeness. Given an original sentence $x$ and its adversarial text $x^*$ generated by either one of the attacks, we design a human study to \textit{directly compare} {\mymethod} with \textit{TextBugger}. Specifically, two alternative hypotheses for our validation are (1) $\mathcal{H}_\mathrm{Semantics}$: $x^*$ generated by {\mymethod} preserves the original meanings of $x$ \textit{better} than that generated by \textit{TextBugger} and (2) $\mathcal{H}_\mathrm{Human}$: $x^*$ generated by {\mymethod} is \textit{more likely} to be perceived as a human-written text (and not machine) than that generated by \textit{TextBugger}.
\vspace{5pt}

\begin{figure}[t!b]
\centerline{\includegraphics[width=0.47\textwidth]{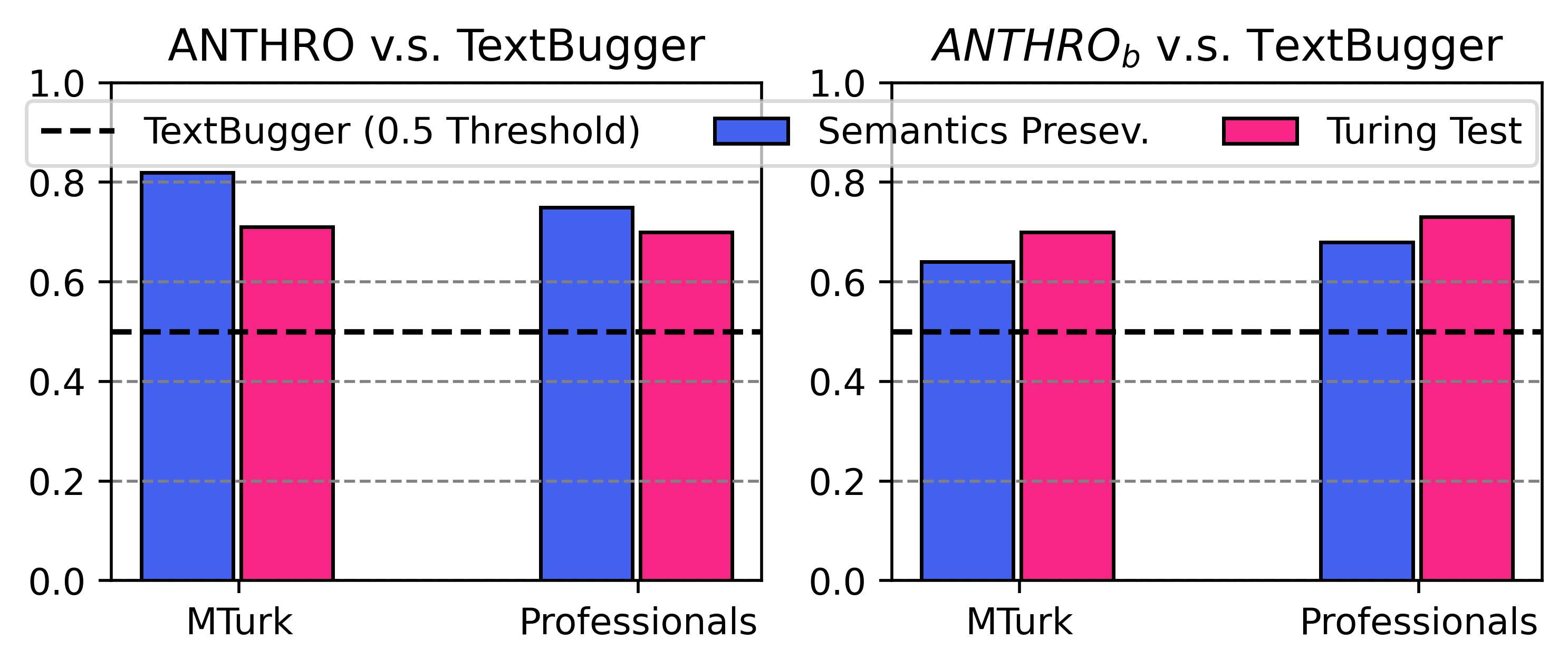}}
\caption{Semantic preservation and human-likeness}
\label{fig:user_study}
% \vspace{-10pt}
\end{figure}

\noindent \textbf{Human Study Design.} We use the two attackers to generate adversarial texts targeting BERT model on 200 examples sampled from the TC dataset's test set. We then gather examples that are successfully attacked by both {\mymethod} and \textit{TextBugger}. Next, we present a pair of texts, one generated by {\mymethod} and one by \textit{TextBugger}, together with the original sentence to human subjects. We then ask them to select (1) which text \textit{better} preserves the meaning of the original sentence (Figure \ref{fig:amt1} in Appendix) and (2) which text is \textit{more likely} to be written by human (Figure \ref{fig:amt2} in Appendix). \textcolor{black}{To reduce noise and bias, we also provide a \textit{``Cannot decide"} option when quality of both texts are equally good or bad, and present the two questions in two separate tasks. Since the definition of semantic preservation can be subjective, we recruit human subjects as both (1) Amazon Mechanical Turk (MTurk) workers and (2) professional data annotators at a company with extended experience in annotating texts in domain such as toxic and hate speech}. Our human subject study with MTurk workers was IRB-approved. \textcolor{black}{We refer the readers to Sec. \ref{appendix:human_study} (Appendix) for more details on MTurks and study designs.}
\vspace{5pt}

\renewcommand{\tabcolsep}{1pt}
\begin{table}[t!b]
    \centering
    \footnotesize
    \begin{tabular}{lHcc}
        \toprule
        \multirow{2}{*}{\textbf{Reason}} & \textbf{Origin} & \textbf{Favorable} & \textbf{Unfavorable} \\
        {} & {} & \textbf{For {\mymethod}} & \textbf{For TextBugger} \\
        \cmidrule(lr){1-4}
        Genuine Typos & stupid, butt, Faggot & stuupid, but, Faoggt & sutpid, burt, Foggat \\
        Intelligible & failure & faiilure & faioure\\
        Sound Preserv. & shitty, crap & shytty, crp & shtty, crsp \\
        Meaning Preserv. & gay, asshole, dumb & ga-y, ashole, dummb & bay, alshose, dub \\
        High Search Results & racist, kills & sodmized, kiills & Smdooized, klils \\
        Personal Exposure & ignorant, garbage & ign0rant, gaarbage & ignorajt, garage\\
        Word Selection & & morons$\rightarrow$mor0ns & edited$\rightarrow$ewited \\
        \bottomrule
    \end{tabular}
    \caption{Top reasons in favoring {\mymethod}'s perturbations as more likely to be written by human.}
    \label{tab:qualitative}
    % \vspace{-15pt}
\end{table}

\begin{figure}[t!b]
% \hspace{-10pt}
\centerline{\includegraphics[width=0.47\textwidth]{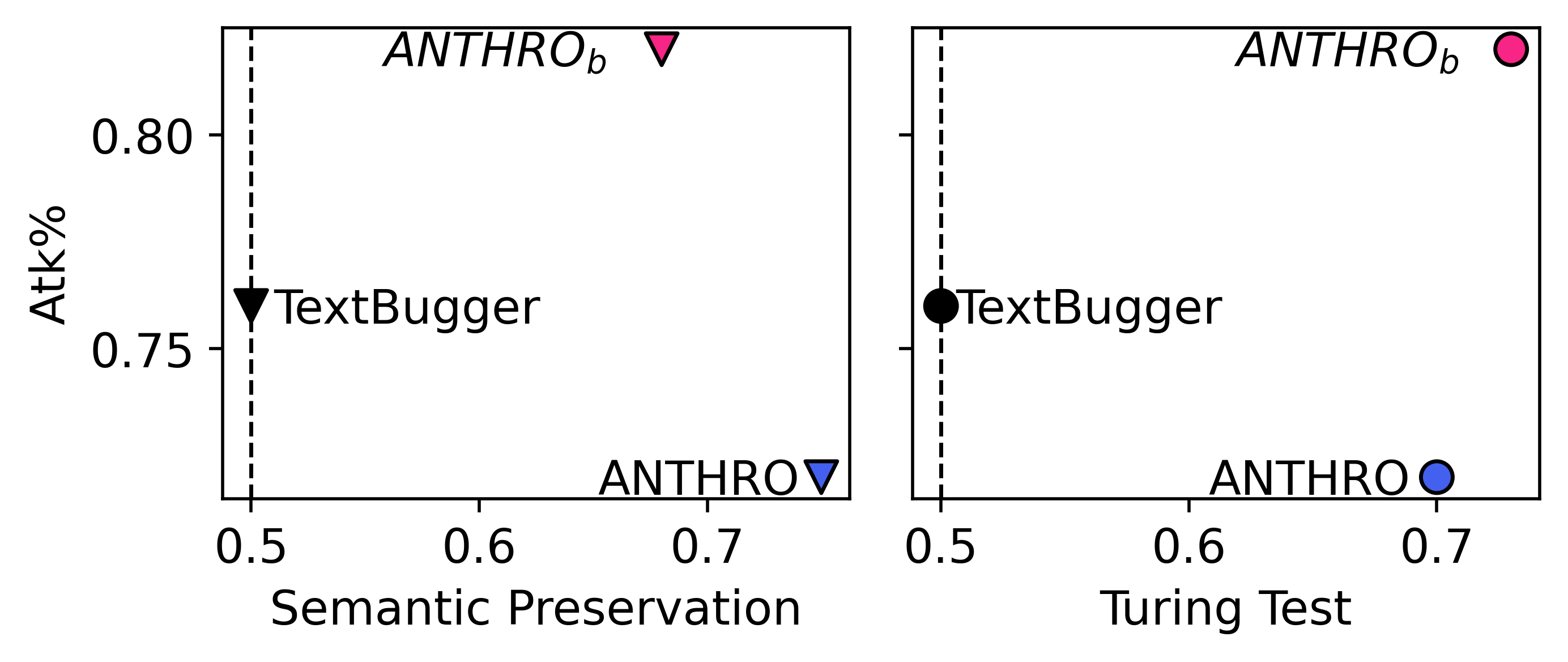}}
\caption{Trade-off among evaluation metrics}
\label{fig:tradeoff}
% \vspace{-10pt}
\end{figure}

\noindent \textbf{Quantitative Results.} It is statistically significant (\textit{p-value}$\leq$0.05) to reject the null hypotheses of both $\mathcal{H}_\mathrm{Semantics}$ and $\mathcal{H}_\mathrm{Human}$ (Table \ref{tab:user_study}). Overall, adversarial texts generated by perturbations mined in the wild are much better at preserving the original semantics and also at resembling human-written texts than those generated by \textit{TextBugger} (Figure \ref{fig:user_study}, Left). 
\vspace{5pt}

\noindent \textbf{Qualitative Analysis.} 
% We also ask the professional subjects to provide optional comments on their thought process. 
Table \ref{tab:qualitative} summarizes the top reasons why they favor {\mymethod} over \textit{TextBugger} in terms of human-likeness. {\mymethod}'s perturbations are perceived similar to genuine typos and more intelligible. They also better preserve both meanings and sounds. Moreover, some annotators also rely on personal exposure on Reddit, YouTube comments, or the frequency of word use via the search function on Reddit to decide if a word-choice is human-written. 
% Interestingly, one mentions that {\mymethod} is better at selecting sensible words--i.e., ``morons" instead of ``edit", to perturb than \textit{TextBugger}, even though the two methods share the same iterative attack mechanism (Alg. \ref{alg:attack}). This happens because {\mymethod} directly ensembles the distribution of human-written texts, which naturally includes more replacement candidates for offensive than non-offensive words. This eventually increases the probability of sensitive words being perturbed. 
% \vspace{5pt}
% \noindent \textbf{Trade-off between Atk\% and Human Evaluation.} Figure \ref{fig:tradeoff} (Appendix) illustrates the trade-off between attack performance (Atk\%), semantic preservation and human-likeness. Overall, {\mymethodx} is best positioned in the top-right corner of the trade-off plots. Compared to \textit{TextBugger}, our proposed attacks consistently make adversarial texts more practical--i.e., meanings are better preserved, and more challenging to be detected as machine-generated even by experienced annotators. This makes {\mymethod} very promising in practice.

\section{{\mymethodx} Attack}\label{sec:anthro_beta}
\noindent \textbf{{\mymethodx}.} We examine if perturbations inductively extracted from the wild help improve the deductive \textit{TextBugger} attack. Hence, we introduce {\mymethodx}, which considers the perturbation candidates from both {\mymethod} and \textit{TextBugger} in Ln. 10 of Alg. \ref{alg:attack}. Alg. \ref{alg:attack} still selects the perturbation that best flip the target model's prediction.
\vspace{5pt}

\noindent \textbf{Attack Performance.} Even though {\mymethod} comes second after \textit{TextBugger} when attacking BERT model, Table \ref{tab:results2} shows that when combined with \textit{TextBugger}--i.e., {\mymethodx}, it consistently achieves superior performance with an average of 82.7\% and 90.7\% in Atk\% on BERT and RoBERTa even under all normalizers (A+H+P).
\vspace{5pt}

\noindent \textbf{Semantic Preservation and Human-Likeness.} {\mymethodx} improves \textit{TextBugger}'s Atk\%, semantic preservation and human-likeness score with an increase of over 8\%, 32\% and 42\% (from 0.5 threshold) on average (Table \ref{tab:results2},  \ref{fig:user_study}, Right), respectively. The presence of only a few human-like perturbations generated by {\mymethod} is sufficient to signal whether or not the whole sentence is written by humans, while only one unreasonable perturbation generated by \textit{TextBugger} can adversely affect its meaning. This explains the performance drop in terms of semantic preservation but not in human-likeness when indirectly comparing {\mymethodx} with {\mymethod}. Overall, {\mymethodx} also has the best trade-off between Atk\% and human evaluation--i.e., positioning at top right corners in Figure \ref{fig:tradeoff}, with a noticeable superior Atk\%.
% Particularly, {\mymethodx} trade-offs from {\mymethod} some reduction in semantic preservation for superior Atk\%. This gain iterates the overall benefits of human-written perturbations for adversarial attacks.
% Human-written perturbations not only help improve semantic preservation and human-likeness but also Atk\% when augmented with a deductive attack such as \textit{TextBugger}.

\renewcommand{\tabcolsep}{4pt}
\begin{table}[tb]
    \centering
    \footnotesize
    \begin{tabular}{lcccccc}
    \toprule
        \textbf{Model} & \multicolumn{3}{c}{{\textbf{\mymethod}}} & \multicolumn{3}{c}{{\textbf{$\mathrm{{\mymethod}_\beta}$}}} \\
        \cmidrule(lr){2-4}\cmidrule{5-7}
        & \textbf{TC}$\downarrow$ & \textbf{HS}$\downarrow$ & \textbf{CB}$\downarrow$ & \textbf{TC}$\downarrow$ & \textbf{HS}$\downarrow$ & \textbf{CB}$\downarrow$ \\
        \cmidrule(lr){1-7}
        BERT & 0.72 & 0.82 & 0.71 & \uline{0.82} & 0.97 & 0.88\\
        % RoBERTa & 0.84 & 0.93 & 0.78 & 0.91 & 0.97 & 0.89\\
        BERT+A+H+P & 0.65 & 0.65 & 0.60 & 0.85 & \uline{0.79} & \uline{0.84}\\
        % RoBERTa+3Norm &0.80 & 0.91 & 0.82 & 0.88 & 0.93 & 0.91\\
        \cmidrule(lr){1-7}
        % BERT+Ext.Norm & \\
        \textsc{Adv.Train} & \uline{0.41} & \uline{0.30}  & \uline{0.35} & \textbf{0.72} & \textbf{0.72} & \textbf{0.67}\\
        \textsc{SoundCNN} & \textbf{0.14} & \textbf{0.02} & \textbf{0.15} & 0.86& 0.84& 0.92\\
        % \textsc{BERT+SoundCNN} & 0.40& 0.41& 0.40& 0.70& 0.80& 0.72\\
    \bottomrule
    \end{tabular}
    \caption{Averaged Atk\% of {\mymethod} and {\mymethodx} against different defense models.}
    \label{tab:results_defense}
    % \vspace{-10pt}
\end{table}

\section{Defend {\mymethod}, {\mymethodx} Attack} \label{sec:defense}
We suggest two countermeasures against {\mymethod} attack. They are \textbf{(i) Sound-Invariant Model (\textsc{SoundCNN}):} When the defender do \textit{not} have access to $\{\mathcal{H}\}_0^K$ learned by the attacker, the defender trains a generic model that encodes not the spellings but the phonetic features of a text for prediction. Here we train a CNN model~\cite{kim-2014-convolutional} on top of a embeddings layer for discrete \textsc{Soundex++} encodings of each token in a sentence; \textbf{(ii) Adversarial Training (\textsc{Adv.Train}):} To overcome the lack of access to $\{\mathcal{H}\}_0^K$, the defender extracts his/her perturbations in the wild from a separate corpus $\mathcal{D}^*$ where $\mathcal{D}^*{\cap}\mathcal{D}{=}{\emptyset}$ and uses them to augment the training examples--i.e., via self-attack with ratio 1:1, to fine-tune a more robust BERT model. We use $\mathcal{D}^*$ as a corpus of 34M general comments from online news. We compare the two defenses against BERT and BERT combined with 3 layers of normalization A+H+P. BERT is selected as it is better than RoBERTa at defending against {\mymethod} (Table \ref{tab:results}). 
\vspace{5pt}

\noindent \textbf{Results.} Table \ref{tab:results_defense} shows that both \textsc{SoundCNN} and \textsc{Adv.Train} are robust against {\mymethod} attack, while \textsc{Adv.Train} performs best when defending {\mymethodx}. Since \textsc{SoundCNN} is strictly based on phonetic features, it is vulnerable against {\mymethodx} whenever \textit{TextBugger}'s perturbations are selected. Table \ref{tab:results_defense} also underscores that {\mymethodx} is a strong and practical attack, defense against which is thus an important future direction.

\section{Discussion and Analysis}\label{sec:discussion}

\begin{figure}[t!b]
\hspace{-5pt}
% \vspace{-15pt}
\centerline{\includegraphics[width=0.49\textwidth]{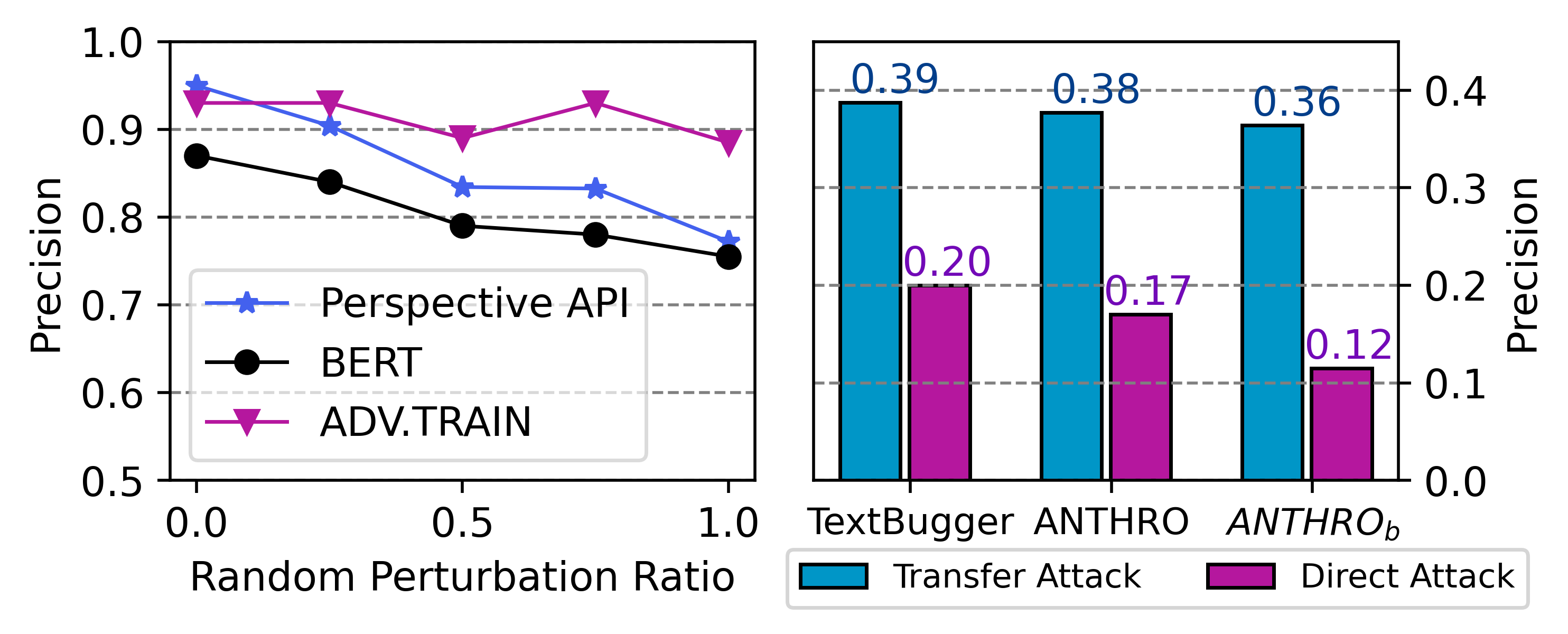}}
\caption{(Left) Precision on human-written perturbed texts synthesized by {\mymethod} and (Right) Robustness evaluation of \textit{Perspective API} under different attacks}
\label{fig:realistic}
% \vspace{-10pt}
\end{figure}

\noindent \textbf{Evaluation with \textit{Perspective API}.} 
% The understanding of different variations of human-written texts is critical to fully capture the semantic meanings of inputs especially in sensitive domains such as toxicity moderation. 
We evaluate if {\mymethod} and {\mymethodx} can successfully attack the popular \textit{Perspective API}~\footnote{\url{https://www.perspectiveapi.com/}}, which has been adopted in various publishers--e.g., NYTimes, and platforms--e.g., Disqus, Reddit, to detect toxicity. We evaluate on 200 toxic texts randomly sampled from the TC dataset.
% We evaluate (1) if the API can capture different forms of human-written toxic texts by using {\mymethod} to randomly perturb different portions of words in 200 positive texts from the TC dataset, (2) if the API can defend against {\mymethod} attack either via a direct iteration mechanism (Alg. \ref{alg:attack}) or via transfer attack through an intermediate BERT classifier. 
Figure \ref{fig:realistic} (Left) shows that the API provides superior performance compared to a self fine-tuned BERT classifier, yet its precision deteriorates quickly from 0.95 to only 0.9 and 0.82 when 25\%--50\% of a sentence are randomly perturbed using human-written perturbations. However, the \textsc{Adv.Train} (Sec. \ref{sec:defense}) model achieves fairly consistent precision in the same setting. This shows that {\mymethod} is not only a powerful and realistic attack, but also can help develop more robust text classifiers in practice. The API is also vulnerable against both direct (Alg. \ref{alg:attack}) and transfer {\mymethod} attacks through an intermediate BERT classifier, with its precision dropped to only 0.12 when evaluated against {\mymethodx}. 
\vspace{5pt}

\renewcommand{\tabcolsep}{1.5pt}
\begin{table}[tb]
    \centering
    \begin{tabular}{ccc}
        \toprule
        \textbf{Task} & \textbf{Sentiment Analysis} & \textbf{Categorization} \\
        % \cmidrule(lr){1-3}
        % Clean & 0.91\\
        \cmidrule(lr){1-3}
        {\mymethod} & 0.80 & 0.93\\
        {\mymethodx} & 0.86 & 1.00 \\
        \bottomrule
    \end{tabular}
    \caption{Averaged Atk\% of {\mymethod} and {\mymethodx} in fooling Google Cloud\footnote{https://cloud.google.com/natural-language}'s sentiment analysis API and text categorization API.}
    \label{tab:non_abusive}
\end{table}

\noindent \textcolor{black}{\textbf{Generalization beyond Offensive Texts.} Although {\mymethod} extracts perturbations from abusive data, the majority of them are non-abusive texts. Thus, {\mymethod} learns perturbations for non-abusive English words--e.g., hilarious->Hi-Larious, shot->sh•t. We also make no assumption on the task domains that {\mymethod} can attack. Evidently, {\mymethod} and {\mymethodx} achieves 80\%, 86\% Atk\% and 90\%, 100\% Atk\% on fooling the sentiment analysis and text categorization API from Google Cloud (Table \ref{tab:non_abusive})}
\vspace{5pt}

\noindent \textcolor{black}{\textbf{Computational Complexity.}} \label{sec:complexity}
The \textbf{one-time} extraction of $\{\mathcal{H}\}_0^K$ via Eq. (\ref{eqn:extraction}) has $\mathcal{O}(|\mathcal{D}|L)$ where $|\mathcal{D}|$, $L$ is the \# of tokens and the length of longest token in $\mathcal{D}$ (hash-map operations cost $\mathcal{O}(1)$). Given a word $w$ and $\mathbf{k},\mathbf{d}$, {\mymethod} retrieves a list of perturbation candidates via Eq. (\ref{eqn:searching}) with $\mathcal{O}(|w|max(\mathcal{H}_k))$ where $|w|$ is the length of $w$ and $max(\mathcal{H}_k)$ is the size of the largest set of tokens sharing the same \textsc{Soundex++} encoding in $\mathcal{H}_k$. Since $max(\mathcal{H}_k)$ is constant, the upper-bound then becomes $\mathcal{O}(|w|)$.
\vspace{5pt}

\noindent \textcolor{black}{\textbf{Limitation of Misspelling Correctors.} Similar to other spell-checkers such as \textit{pyspellchecker} and \textit{symspell}, the SOTA NeuSpell depends on a fixed dictionary of common misspellings, or synthetic misspellings generated by random permutation of characters~\cite{neuspell}. These checkers often assume perturbations are within an edit-distance threshold from the original words. This makes them exclusive since one can easily generate new perturbations by repeating a specific character--e.g., ``porn"$\rightarrow$``pooorn". Also, due to the iterative attack mechanism (Alg. \ref{alg:attack}) where each token in a sentence is replaced by many candidates until the correct label's prediction probability drops, {\mymethod} only needs a single good perturbation that is not detected by NeuSpell for a successful replacement. Thus, by formulating perturbations by not only their spellings but also their sounds, ANTHRO is able to mine perturbations that can circumvent NeuSpell.}
\vspace{5pt}

\noindent \textbf{Limitation of {\mymethod}.} The perturbation candidate retrieval operation (Eq. (\ref{eqn:searching})) has a higher computational complexity than that of other methods--i.e., $\mathcal{O}(|w|)$ v.s. $\mathcal{O}(1)$ where $|w|$ is the length of an input token $w$ (Please refer to Sec. \ref{sec:complexity} in the Appendix for detailed computational complexity). This can prolong the running time, especially when attacking long documents. However, we can overcome this by storing all the perturbations (given $\mathbf{k},\mathbf{d}$) of the top frequently used offensive and non-offensive English words. We can then expect the operation to have an average complexity close to $\mathcal{O}(1)$. The current \textsc{Soundex++} algorithm is designed for English texts and might not be applicable in other languages. Thus, we plan to extend {\mymethod} to a multilingual setting.

\section{Conclusion}

We propose {\mymethod}, a character-based attack algorithm that extracts human-written perturbations in the wild and then utilizes them for adversarial text generation. Our approach yields the best trade-off between attack performance, semantic preservation and stealthiness under both empirical experiments and human studies. A BERT classifier trained with examples augmented by {\mymethod} can also better understand human-written texts. 
% \textcolor{black}{Our work also opens a new direction of research on how netizens utilize text perturbations for censorship in this new age of AI.}
% \lee{reference is not properly sorted??}
% \section{References}
% \newpage
% \clearpage
\section*{Broad Impact}
To the best of our knowledge, {\mymethod} is the first work that extracts noisy human-written texts, or text perturbations, online. We further iterate what reviewer pvcD has observed: {\mymethod} moves ``away from deductively-derived attacks to data-driven inspired attacks". This novel direction is beneficial not only to the adversarial NLP community but also in other NLP tasks that require the understanding of realistic noisy user-generated texts online. Specifically, Sec. \ref{sec:defense} and Figure \ref{fig:realistic} shows that our work enables the training of a BERT model that can understand noisy human-written texts better than the popular \textit{Perspective API}. By extending this to other NLP tasks such as Q\&A and NLI, our work hopes to enable current NLP software to perform well in real life settings, especially on social platforms where user-generated texts are not always in perfect English. Our work also opens a new direction in  the use of languages online and how netizens utilize different forms of perturbations for avoiding censorship in this new age of AI.

\section*{Ethical Consideration}
Similar to previous works in adversarial NLP literature, there are risks that our proposed approach may be unintentionally utilized by malicious actors to attack textual ML systems. To mitigate this, we will not publicly release the full perturbation dictionary that we have extracted and reported in the paper. Instead, we will provide access to our private API on a case-by-case basis with proper security measures. Moreover, we also suggest and discuss two potential approaches that can defend against our proposed attacks (Sec. \ref{sec:defense}). We believe that the benefits of our work overweight its potential risks. All public secondary datasets used in this paper were either open-sourced or released by the original authors. 

\section*{Acknowledgement}
This research was supported in part by NSF awards \#1820609, \#1915801, and \#2114824. 

\newpage
\clearpage
\bibliography{custom}
\bibliographystyle{acl_natbib}

\newpage
\clearpage
\appendix
\section{Supplementary Materials}
\setcounter{table}{0}
\setcounter{figure}{0}
\renewcommand\thetable{\Alph{section}.\arabic{table}}
\renewcommand\thefigure{\Alph{section}.\arabic{figure}}

\subsection{Additional Results and Figures}
Below are list of supplementary materials:
\begin{itemize}
    \item Table \ref{tab:dataset}: list of datasets we used to curate the corpus $\mathcal{D}$, from which human-written perturbations are extracted (Sec. \ref{sec:mining_perturbations}). All the datasets are publicly available, except from the two private datasets \textit{Sensitive Query} and \textit{Hateful Comments}. 
    \item Table \ref{tab:attack_dataset}: list of datasets we used to evaluate the attack performance of all attackers (Sec. \ref{sec:attack_performance}) and the prediction performance of BERT and RoBERTa on the respective test sets. All datasets are publicly available.
    \item Table \ref{tab:user_study}: Statistical analysis of the human study results (Sec. \ref{sec:human_study}). 
    % \item Table \ref{tab:qualitative}: List of top reasons provided by the professional annotators on why they prefer {\mymethod} over \textit{TextBugger} in the human-likeness test (Sec. \ref{sec:human_study}).
    \item Figure \ref{fig:wordcloud}: Word-cloud of extracted human-written perturbations by {\mymethod} for some of popular English words.
    \item Figure \ref{fig:amt1}, \ref{fig:amt2}: Interfaces of the human study described in Sec. \ref{sec:human_study}.
\end{itemize}

\subsection{Infrastructure and Software}

% \begin{figure}[t!]
% % https://app.diagrams.net/#G14BI6M-am1gCjrpIDbfuEVgMLhmZYx_2x
% \centerline{\includegraphics[width=0.47\textwidth]{figures/wordcloud_democrats_republicans3.png}}
% \caption{Word-clouds of human-written perturbations for the English word ``democrats" and ``republicans"}
% \label{fig:wordcloud}
% \vspace{-10pt}
% \end{figure}

\renewcommand{\tabcolsep}{0.7pt}
 \begin{table}[t]
 \centering
 \small
 \begin{tabular}{lHccHHH}
    \toprule
    \multirow{2}{*}{\textbf{Dataset}}  & \multirow{2}{*}{\textbf{Source}} & \multirow{2}{*}{\textbf{\#Texts}}  & \multirow{2}{*}{\textbf{\#Tokens}} & \textbf{\#Sounds} & \textbf{\#Avg.Pert.} & \textbf{\#Avg.Pert.}\\
    {} & {} & {} & {} &  \textbf{(level 1)} & \textbf{(case-insensitive)} & \textbf{(case-sensitive)} \\
    \cmidrule{1-7}
    List of Bad Words~\footnote{\url{https://github.com/zacanger/profane-words}} & & 1.9K & 1.9K\\
     Rumours (Twitter)~\cite{kochkina2018all} & Twitter & 99K  & 159K & 69K & 3.07 & 10.28\\
     Hate Memes (Twitter)~\cite{gomez2020exploring} & Twitter & 150K & 328K\\
     Personal Atks (Wiki.)~\cite{wulczyn_thain_dixon_2017} & Wikipedia & 116K & 454K \\
     Toxic Comments (Wiki.) (Kaggle, 2019) & Wikipedia & 2M  & 1.6M & 441K & 4.39 & 29.75\\
     Malignant Texts (Reddit) (Kaggle, 2021)\footnote{https://www.kaggle.com/surekharamireddy/malignant-comment-classification} & Unknown & 313K &  857K\\
     Hateful Comments (Reddit) (Kaggle, 2021)\footnote{https://www.kaggle.com/jholger/hateful-and-controversial-reddit-comments} & Reddit & 1.7M & 1M\\
     \cmidrule(lr){1-7}
     Sensitive Query (Search Engine, Private) & Search Eng. & 1.2M  & 314K & 82K & 1.26 & 26.81\\
     Hateful Comments (Online News, Private) & Online News & 12.7M  & 7M & 1.7M & 3.127 & 13.27\\
     \cmidrule{1-7}
     \textbf{Total texts used to extract {\mymethod}} & & \textbf{18.3M} & -\\
    \bottomrule
     \end{tabular}
    \caption{Real-life datasets that are used to extract adversarial texts in the wild, number of total examples (\#Texts) and unique tokens (\#Tokens) (case-insensitive)}
     \label{tab:dataset}
\end{table}

\begin{table}[tb]
    \centering
    \footnotesize
    \begin{tabular}{lHccc}
    \toprule
        \textbf{Dataset} & \textbf{Source} & \textbf{\#Total} & \textbf{BERT} & \textbf{RoBERTa} \\
        \cmidrule(lr){1-5}
        CB~\cite{cyberbullyingdata} & Multi* & 449K & 0.84 & 0.84\\
        TC (Kaggle, 2018) & Wikipedia & 160K & 0.85 & 0.85\\
        % SMS Spam~\cite{Almeida2011SpamFiltering} & Email &  34K & 1.00 & 1.00\\
        HS~\cite{hateoffensive} & Twitter &  25K & 0.91 & 0.97\\
    \bottomrule
    % \multicolumn{4}{l}{*Comments from Kaggle, YouTube, Twitter}
    \end{tabular}
    \caption{Evaluation datasets Cyberbullying (CB), Toxic Comments (TC) and Hate Speech (HS) and prediction performance in F1 score on their test sets of BERT and RoBERTa.}
    \label{tab:attack_dataset}
\end{table}

\renewcommand{\tabcolsep}{1pt}
\begin{table}[t!h]
    \centering
    \footnotesize
    \begin{tabular}{ccccc}
        \toprule
        \textbf{Alternative Hypothesis} & \textbf{Mean} & \textbf{t-stats} & \textbf{p-value} & \textbf{df}\\
        \cmidrule(lr){1-5}
        \multicolumn{5}{c}{----- \uline{AMT Workers} as Subjects -----} \\
        \\
        $\mathcal{H}_{\mathrm{Semantics}}:$ $\mathrm{{\mymethod}}$ > TB & 0.82 & 5.66 & 4.1e-7\textbf{**} & 48\\
        $\mathcal{H}_{\mathrm{Semantics}}:$ $\mathrm{{\mymethod}_\beta}$ > TB & 0.64 & 1.95 & 2.9e-2\textbf{*} & 46\\
        $\mathcal{H}_{\mathrm{Human}}:$ $\mathrm{{\mymethod}}$ > TB & 0.71 & 3.14 & 1.5e-3\textbf{**} & 47\\
        $\mathcal{H}_{\mathrm{Human}}:$ $\mathrm{{\mymethod}_\beta}$ > TB & 0.70 & 3.00 & 2.2e-3\textbf{**} & 46\\
        \cmidrule(lr){1-5}
        \multicolumn{5}{c}{----- \uline{Professional Annotators} as Subjects -----} \\
        \\
        $\mathcal{H}_{\mathrm{Semantics}}:$ $\mathrm{{\mymethod}}$ > TB & 0.75 & 3.79 & 2.4e-4\textbf{**} & 44\\
        $\mathcal{H}_{\mathrm{Semantics}}:$ $\mathrm{{\mymethod}_\beta}$ > TB & 0.68 & 2.49 & 8.6e-3\textbf{**} & 41 \\
        $\mathcal{H}_{\mathrm{Human}}:$ $\mathrm{{\mymethod}}$ > TB & 0.70 & 3.06 & 1.82e-3\textbf{**} & 50\\
        $\mathcal{H}_{\mathrm{Human}}:$ $\mathrm{{\mymethod}_\beta}$ > TB & 0.73 & 3.53 & 4.6e-4\textbf{**} & 48 \\
        % \multicolumn{5}{c}{----- \textit{Without} Perturbation Highlighting -----} \\
        % $\mathcal{H}_{\mathrm{Semantics}}:$ ANTHRO > TB & 0.77 & 4.42 & 2.9e-5\textbf{**} & 47\\
        % $\mathcal{H}_{\mathrm{Turing}}:$ ANTHRO > TB & 0.74 & 3.65 & 3.4e-4\textbf{**} & 45\\
        \bottomrule
        \multicolumn{5}{l}{Statistical significant \textbf{**}(p-value$\leq$0.01) \textbf{*}(p-value$\leq$0.05)}
    \end{tabular}
    \caption{It is \textit{statistically significant} (p-value$\leq$0.01) that adversarial texts generated by {\mymethod} are better than those generated by TextBugger (TB) at both preserving the semantics of the original sentences ($\mathcal{H}_\mathrm{Semantics}$)) and at being perceived as human-written texts ($\mathcal{H}_\mathrm{Human}$).}
    \label{tab:user_study}
\end{table}

\section{Implementation Details}
\setcounter{table}{0}
\setcounter{figure}{0}
\renewcommand\thetable{\Alph{section}.\arabic{table}}
\renewcommand\thefigure{\Alph{section}.\arabic{figure}}

\subsection{Attackers}
We evaluate all the attack baselines using the open-source \textit{OpenAttack} framework~\cite{zeng2020openattack}. We keep all the default parameters for all the attack methods. 

\subsection{Defenders}
For the (1) \textit{Accents normalization}, we adopt the accents removal code from the \textit{Hugging Face} repository~\footnote{\url{https://huggingface.co}}. For (2) \textit{Homoglyph normalization}, we adopt a 3rd party python \textit{Homoglyph} library\footnote{\url{https://github.com/codebox/homoglyph}}. For (3) \textit{Perturbation normalization}, we use the state-of-the-art misspelling-based perturbation correction \textit{Neuspell} model~\cite{neuspell}~\footnote{\url{https://github.com/neuspell/neuspell}}. For \textit{Perspective API}, we directly use the publicly available API provided by Jigsaw and Google~\footnote{\url{https://www.perspectiveapi.com/}}.

\subsection{Details of Human Study and Experiment Controls}\label{appendix:human_study}
To ensure a high quality response from MTurks, we require a minimum attentions span of 30 seconds for each question. We recruit MTurk workers who are 18 years or older residing in North America. MTurk workers are recruited using the following qualifications provided by AMT, namely (1) recognized as ``master'' workers by AMT system, (2) have done at least 5K HITs and (3) have historical HITs approval rate of at least 98\%. These qualifications are also more conservative than previous human studies we found in previous literature. We pay each worker on average around \$10 an hour or higher (federal minimum wage was \$7.25 in 2021 when we carried out our study). To limit abusive behaviors, we impose a minimum attention span of 30 seconds for the workers to complete each task.

\newpage
\clearpage

\begin{figure*}[t!b]
\centerline{\includegraphics[width=0.8\textwidth]{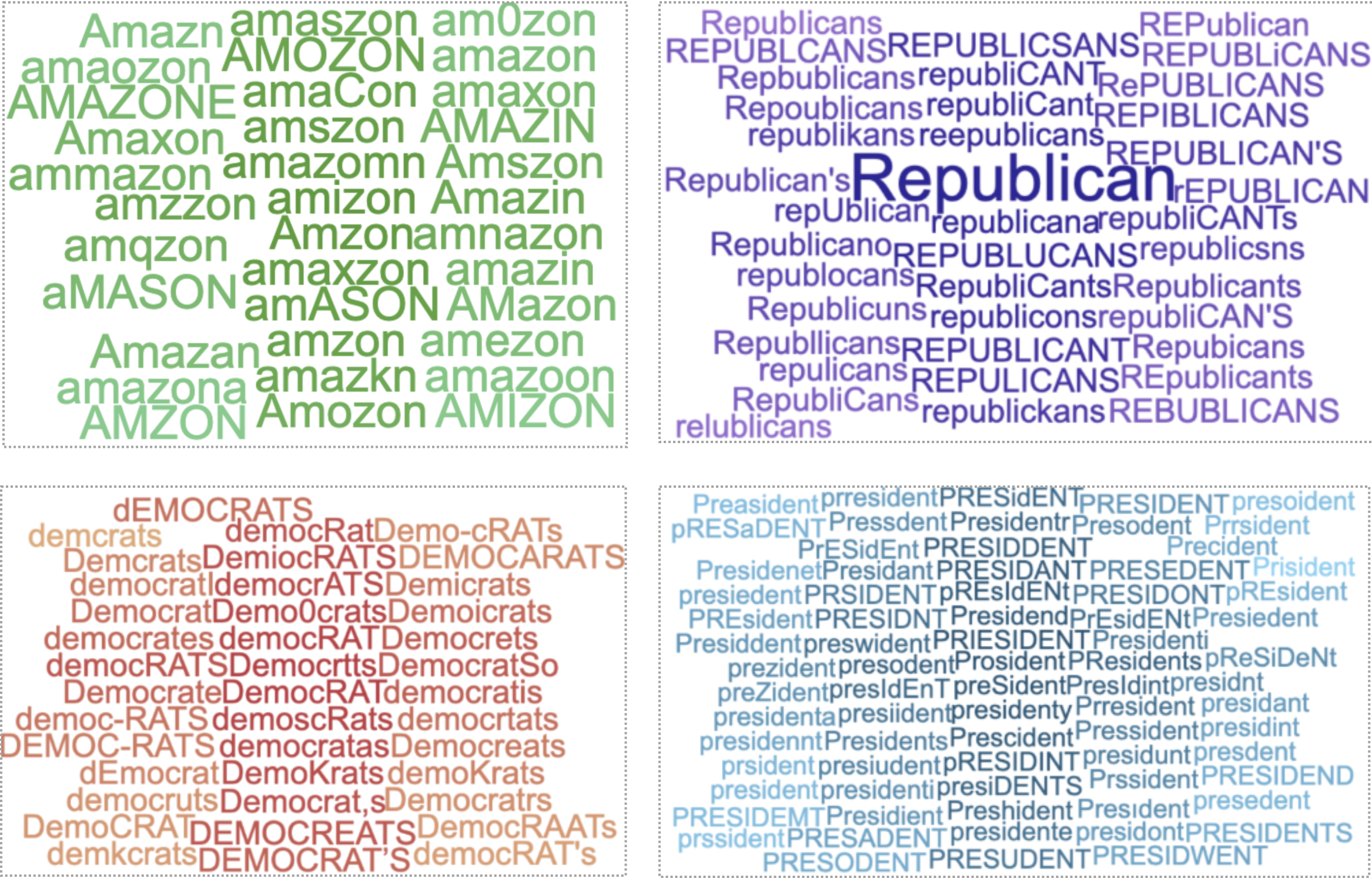}}
\caption{Word-clouds of perturbations in the wild extracted by {\mymethod} for the word ``amazon'', ``republicans'', ``democrats'' and ``president''.}
\label{fig:wordcloud}
\end{figure*}

\begin{figure*}[tb]
\centerline{\includegraphics[width=0.9\textwidth]{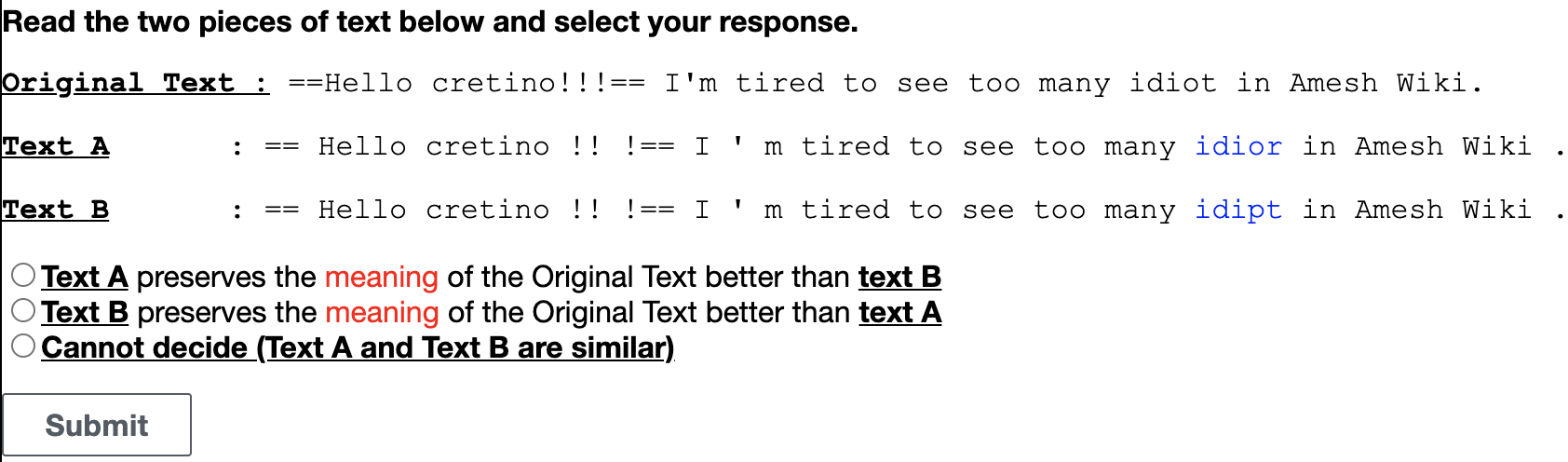}}
\caption{User-study design for semantic preservation comparison between {\mymethod}, {\mymethodx} v.s. \textit{TextBugger}}
\label{fig:amt1}
\end{figure*}

\begin{figure*}[tb]
\centerline{\includegraphics[width=0.87\textwidth]{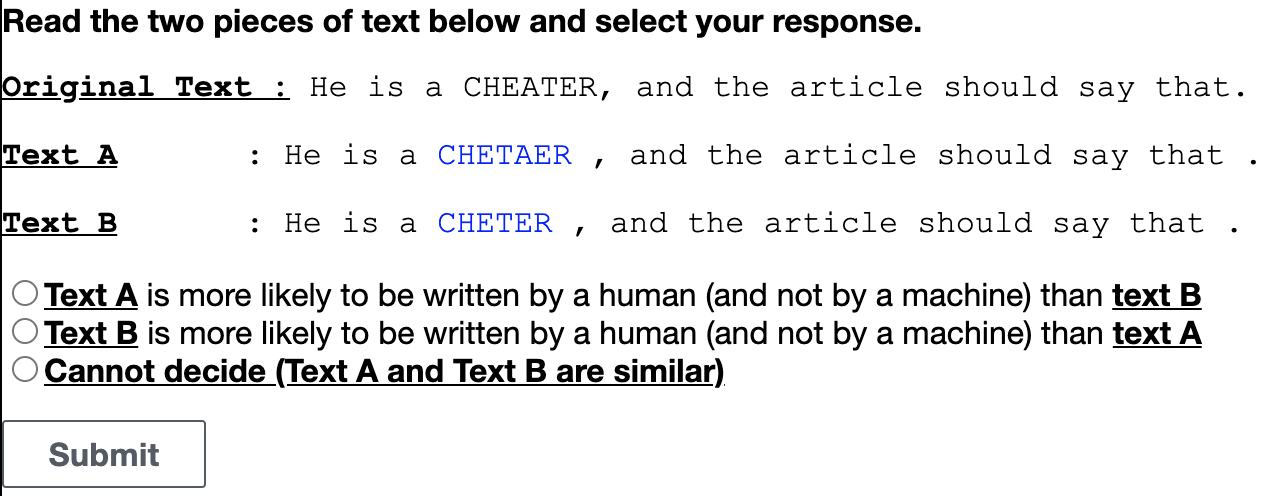}}
\caption{User-study design for human-likeness comparison between {\mymethod}, {\mymethodx} v.s. \textit{TextBugger}}
\label{fig:amt2}
\end{figure*}

\end{document}